\documentclass[letterpaper,twocolumn,10pt]{article}
\usepackage{usenix,epsfig,endnotes}
\usepackage{xspace}
\usepackage{xcolor}
\usepackage{subcaption}
\usepackage{enumitem}
\usepackage{amsmath}
\usepackage{array}
\usepackage{multirow}
\usepackage{tcolorbox}
\usepackage{graphicx}
\usepackage{tikz}
\usepackage{caption}

\begin{document}

\def\sysname{\textsc{HaluProbe}\xspace}
\newcommand{\sssec}[1]{\vspace*{0.05in}\noindent\textbf{#1}}
\newcolumntype{M}[1]{>{\centering\arraybackslash}p{#1}}
\newcommand{\prc}[1]{{\textcolor{red}{[\textit{Peiran: #1}]}}}

%don't want date printed
\date{}

%for single author (just remove % characters)
\author{
{\rm Peiran Wang}\\
ByteDance Inc.
\and
{\rm Yang Liu}\\
ByteDance Inc.
\and
{\rm Yunfei Lu}\\
ByteDance Inc.
\and
{\rm Jue Hong}\\
ByteDance Inc.
\and
{\rm Ye Wu}\\
ByteDance Inc.
% copy the following lines to add more authors
% \and
% {\rm Name}\\
%Name Institution
} % end author

%make title bold and 14 pt font (Latex default is non-bold, 16 pt)
\title{
What are Models Thinking about? Understanding Large Language Model Hallucinations through Model Internal State Analysis
}

\maketitle

\subsection*{Abstract}
Large language model (LLM) systems suffer from the models' unstable ability to generate valid and factual content, resulting in hallucination generation.
Current hallucination detection methods heavily rely on out-of-model information sources, such as RAG to assist the detection, thus bringing heavy additional latency.
Recently, internal states of LLMs' inference have been widely used in numerous research works, such as prompt injection detection, etc.
Considering the interpretability of LLM internal states and the fact that they do not require external information sources, we introduce such states into LLM hallucination detection.
In this paper, we systematically analyze different internal states' revealing features during inference forward and comprehensively evaluate their ability in hallucination detection.
Specifically, we cut the forward process of a large language model into three stages: understanding, query, generation, and extracting the internal state from these stages.
By analyzing these states, we provide a deep understanding of why the hallucinated content is generated and what happened in the internal state of the models.
Then, we introduce these internal states into hallucination detection and conduct comprehensive experiments to discuss the advantages and limitations.
%\footnote{We will release our code upon acceptance.}

% Use the following at camera-ready time to suppress page numbers.
% Comment it out when you first submit the paper for review.
%\thispagestyle{empty}

\section{Introduction}\label{sec:intro}

%\begin{figure}[htbp]
%    \centering
%    \includegraphics[width=0.5\textwidth]{figures/fig-example.pdf}
%    \caption{An example of \sysname's whole process.}
%    \label{fig:example}
%\end{figure}

%The large language model has been widely used in many areas. But people lack trust in it due to hallucinations

Large language models (LLM) have been widely adopted across various fields, ranging from natural language processing and generation tasks to specialized applications in healthcare \cite{yang2022large, thirunavukarasu2023large, peng2023study}, finance \cite{wu2023bloomberggpt}, and legal services \cite{cui2023chatlaw, guha2024legalbench}. 
However, despite its impressive capabilities, a fundamental issue remains: users often lack trust in LLMs due to their tendency to produce ``hallucinations'' \cite{huang2023survey, li2024dawn}.
Hallucination in the context of LLMs refers to the phenomenon where the model generates content that appears plausible but is factually inaccurate or misaligned with the provided context. 
This can result in responses that sound coherent and authoritative, yet introduce misleading or completely false information \cite{mckenna2023sources}.

%Current works towards the hallucination is using some outside knowlegde to intervene in it. However these works suffer from xxxx

Traditional hallucination detection methods usually focus on post-processing analysis of generated outputs, including checking the factual accuracy of responses through external RAGs \cite{gao2023retrieval, niu2023ragtruth, belyi2024luna, li2024enhancing}, adding prefix/suffix/system prompts according to prompt engineering \cite{hanna2023comparative}, and self-consistency checking by generating multiple outputs \cite{harrington2024mitigating, manakul2023selfcheckgpt, mundler2023self}.
However, these methods have some inherent limitations. The reliance on external databases will introduce additional complexity and computational overhead, especially when a large-scale external knowledge base is required \cite{belyi2024luna}. 
Furthermore, the external base may not always be up-to-date or comprehensive, limiting their effectiveness in detecting certain hallucinations.
Also, the prompt engineering-based methods will introduce additional prompts, which may lead to a decrease in response quality \cite{hanna2023comparative}.
At last, since these methods all use external methods for post-generation intervention, there is a lack of understanding of the source of hallucinations.

Recent studies have begun exploring the use of LLM's internal state for hallucination detection and intervention to overcome the limitations of hallucination detection and intervention methods based on external bases.
The internal state of LLM, including attention weights \cite{beigi2024internalinspector, chuang2024lookback, yuksekgonul2023attention}, layer representation \cite{beigi2024internalinspector, ji2024llm, chen2024context, chen2024inside, su2024unsupervised, duan2024llms, azaria2023internal, he2024llm}, logits \cite{quevedo2024detecting, he2024llm}, etc., provides a state representation of the model's reasoning process during content generation.
By analyzing these internal states, signs of hallucination can be detected before the content is fully generated, thereby achieving real-time intervention and reducing computational costs.
Since no external intervention is required, this series of methods usually requires low computational overhead while allowing private models to be deployed locally (no cloud third-party services are required).
Most importantly, these methods have become more interpretable, opening up new research ideas.

\begin{figure}[htbp]
    \centering
    \includegraphics[width=0.5\textwidth]{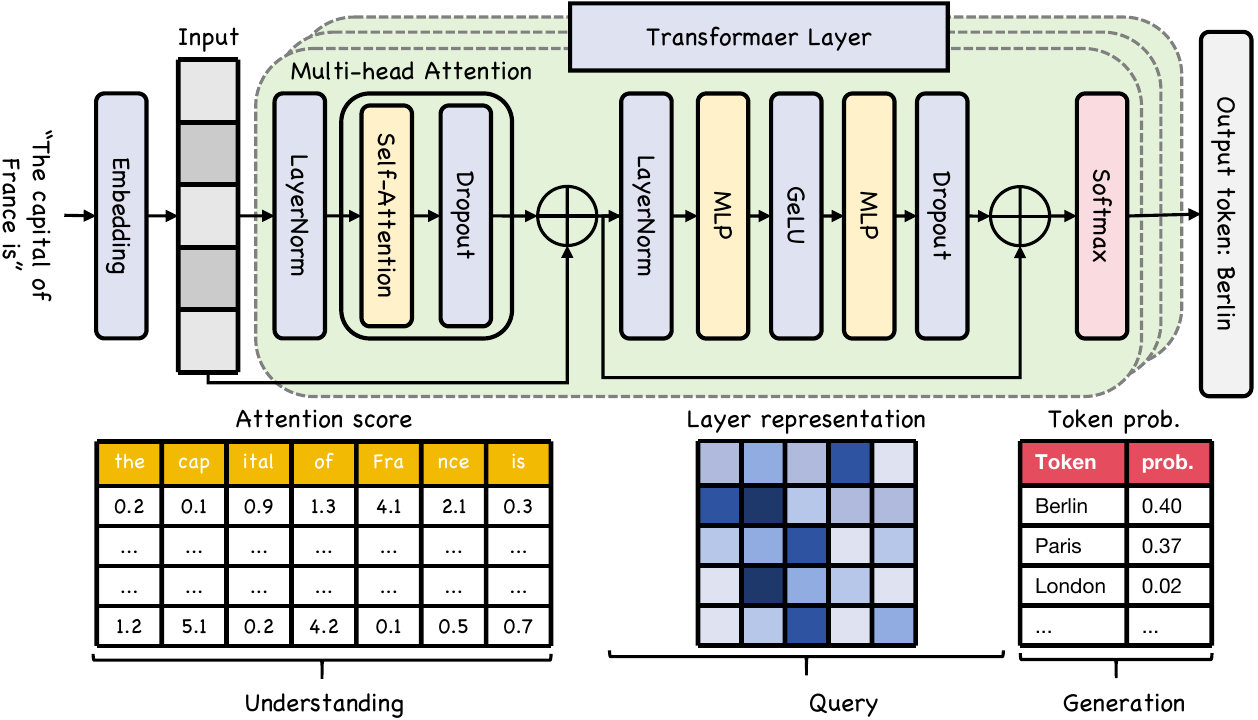}
    \caption{The inference process of LLM can be divided into 4 stages: (1) Input preprocess; (2) Understanding; (3) Query; (4) Generation.}
    \label{fig:infer}
\end{figure}

However, existing work based on internal states focuses on using a certain internal state for detection.
They also did not provide sufficient transferability analysis.
There is a lack of a systematic understanding of how large model hallucinations are generated during the internal inference processes.

To meet this gap, we propose \sysname, a framework that systematically extracts the internal states from the large language model inference and provides a deep understanding of the generation of hallucinations in it.
Specifically, we make three key contributions:
\begin{itemize}[itemsep=0pt, leftmargin=*,topsep=0pt]
    \item We divided LLMs' inference process into three stages: understanding, query, and generation, and extracted 8 features from these three stages.
    \item We did a systematical analysis for all three stages's internal states of inference, including understanding, query, and generation. We provide deep analysis both for solely inference and for inference with RAG.
    \item We conducted comprehensive experiments to assess different features' abilities in hallucination detection. We also considered the transferability across different types of hallucinations.
\end{itemize}

The rest of the paper is organized as follows:
We first introduce the preliminary background knowledge in \S\ref{sec:back}, and summarize current works about hallucination detection and mitigation in \S\ref{sec:related}.
Then we introduce the whole design in \S\ref{sec:scheme}.
Next, we provide a deep understanding and observation about different inner states' modes in hallucinated response in \S\ref{sec:understand}.
The systematical detection performance analysis is stated in \S\ref{sec:detect}.
At last, we discuss the limitation and future work in \S\ref{sec:discussion}.
\section{Preliminary}\label{sec:back}

In this section, we first explore the preliminary definition of LLM hallucination (\S\ref{sec:back:halu}) and the process of LLM inference (\S\ref{sec:back:infer}).

\subsection{LLM Hallucination}\label{sec:back:halu}

\sssec{LLM hallucination definition}.
Hallucination in large language models (LLMs) refers to the generation of content that is inconsistent with reality or factual information \cite{tonmoy2024comprehensive, rawte2023survey, huang2023survey}. 
While this generated content may sound plausible, it is often based on patterns learned from training data rather than actual facts. 
The issue of hallucination significantly impacts the practical applications of LLMs, especially in scenarios requiring high accuracy and reliability, such as healthcare, legal contexts, and journalism \cite{zhang2023siren}.

\sssec{LLM hallucination types}.
Hallucinations in large language models can be categorized into two types:
\begin{enumerate}[label={[\arabic*]}, itemsep=0pt, leftmargin=*,topsep=0pt]
    \item \underline{Input-conflicting hallucination}: This occurs when the content generated by the model deviates from the user’s task instruction or input \cite{tonmoy2024comprehensive, rawte2023survey}. For instance, if a user asks for a dinner recipe, the model might mistakenly provide a lunch suggestion.
    \item \underline{Context-conflicting hallucination}: In multi-turn generation or lengthy content, models may exhibit self-contradictions \cite{tonmoy2024comprehensive, huang2023survey, zhang2023siren}. This type arises when models lose track of the context or fail to maintain consistency throughout the conversation. For example, a model might introduce the current NBA commissioner, Adam Silver, but later refer to a former commissioner, David Stern.
\end{enumerate}

\sssec{Potential causes of LLM hallucination}.
Causes of hallucinations can be analyzed from several perspectives:
\begin{enumerate}[label={[\arabic*]}, itemsep=0pt, leftmargin=*,topsep=0pt]
    \item \underline{Data-related issues}: The training data for LLMs is often collected from the internet, containing significant amounts of outdated, inaccurate, or biased information \cite{tonmoy2024comprehensive, rawte2023survey, zhang2023siren}. These data quality issues can lead to the generation of hallucinated content .
    \item \underline{Limitations of the training process}: Models learn language patterns by predicting the next word rather than understanding the content deeply \cite{rawte2023survey, huang2023survey, ji2023survey}. Additionally, misalignment during fine-tuning (for example, through reinforcement learning from human feedback, or RLHF) can contribute to hallucinations .
    \item \underline{Randomness in the inference stage}: During content generatce inconsistent or contextually irrelevant outputs due to randomness in sampling strategies or decoding methods \cite{tonmoy2024comprehensive, huang2023survey, ji2023survey}.
\end{enumerate}

%\sssec{Dangers of hallucinations}.

%\begin{figure*}[htbp!]
%    \centering
%    \includegraphics[width=\textwidth]{figures/fig-scheme.pdf}
%    \caption{
%    The workflow of \sysname has four stages:
%    (1) Extract internal state from the inference process;
%    (2) Select tokens for hallucination detection;
%    (3) Compute features;
%    (4) Detect hallucination using model.
%    }
%    \label{fig:scheme}
%\end{figure*}

\subsection{LLM Inference Process}\label{sec:back:infer}

During the inference process of a large-scale Transformer model such as Llama, the whole process can be divided into several key stages (see Figure \ref{fig:infer}).

\sssec{Input processing}. The first is the input processing stage, in which the input text is tokenized and converted into a format that the model can understand, generating corresponding token IDs \cite{mickus2022dissect}. Subsequently, these token IDs are input into the embedding layer to form word embeddings, which provide the basis for the model's subsequent understanding. In this way, the text information is effectively converted into numerical representations for subsequent calculations \cite{dar2022analyzing}.

\sssec{Understanding}. The next understanding stage processes the input tokens through a multi-head self-attention mechanism \cite{vaswani2017attention, chefer2021transformer, vig2019analyzing}. This mechanism calculates the similarity between tokens and generates contextual representations to capture long-range dependencies \cite{yeh2023attentionviz}. This process enables the model to fully understand the semantics of each token in a specific context and provide rich semantic information for the subsequent query stage \cite{rigotti2021attention, chefer2021transformer}.

\sssec{Query}. In the query stage, the model further processes the contextual representations generated in the understanding stage through a feedforward network \cite{zhao2024explainability, meng2023locating}. The feedforward network contains two linear transformations and nonlinear activation functions (such as ReLU) to generate new feature representations. The goal of this stage is to extract higher-level abstract information, improve the model's expressiveness, and lay the foundation for the final generation process.

\sssec{Generation}. The generation stage is the core of reasoning. The model converts the output of the query stage into the probability distribution of the next token through linear transformation and softmax function. In this process, the model uses contextual information to determine the most likely next token and generates a continuous text sequence. The decoding strategy of this stage, such as greedy search or beam search, directly affects the quality and coherence of the generated text.

\sssec{Post processing}. Finally, in the post-processing stage, the generated token ID sequence is converted back to a readable text format. By merging subwords or words and removing special tokens, the output of the model is converted into human-understandable language. This stage ensures the readability of the generated content and enables users to interact directly with the model output. Through the close combination of these stages, the Transformer model realizes an effective reasoning process from input to output.
\section{Proposed Scheme: \sysname}\label{sec:scheme}

Our proposed scheme consists of three parts:
(1) Internal state extraction (\S\ref{sec:scheme:state_extract}): \sysname extracts internal states during the inference, and stores them locally.
(2) Token selection (\S\ref{sec:scheme:token_select}): \sysname selects target tokens for next stage feature extraction.
(3) Feature extraction (\S\ref{sec:scheme:feature_extract}): \sysname extracts target features from the internal states, we also discuss different feature selection and token selection for better analysis.
%(4) Detection (\S\ref{sec:detect}): we discuss how to build a detection model using these features.

\subsection{Internal State Extraction}\label{sec:scheme:state_extract}

Following the definition of \S\ref{sec:back:infer}, \sysname first extracts three types of internal state from the three stages during inference:

\sssec{Understanding: attention}.
Attention is the internal state in which LLMs understand the input context.
We extract each attention score matrix from each transformer layer's head for each token $t$ as:
\begin{equation}
    \left[ \alpha_{l,h,i} \right]_{L \times H \times t}=\begin{bmatrix}
    \alpha_{1,1,i} & \alpha_{1,2,i} & \dots & \alpha_{1,H,i} \\
    \alpha_{2,1,i} & \alpha_{2,2,i} & \dots & \alpha_{2,H,i} \\
    \vdots & \vdots & \ddots & \vdots \\
    \alpha_{L,1,i} & \alpha_{L,2,i} & \dots & \alpha_{L,H,i}
    \end{bmatrix}\times t
\end{equation}
where ${\alpha}_{l,h,i}$ denotes the $l$-th layer's $h$-th head's attention score for token $t$ on token $i$.

\sssec{Query: layer representation}.
The query stage involves processing contextual representations through a feedforward network to derive higher-level abstractions. We capture the layer representation matrices from each transformer layer for each token \( t \) as follows:
\begin{equation}
    \left[ \gamma _{l,i} \right]_{L \times t}=\begin{bmatrix}
    \gamma _{1,i} \\
    \gamma _{2,i} \\
    \vdots \\
    \gamma _{L,i}
    \end{bmatrix}\times t
\end{equation}
where \( \gamma _{l,i} \) represents the \( l \)-th layer's layer representation for token \( t \) during the query phase.

\sssec{Generation: logit}.
In the generation stage, the model transforms layer representations into logits, converting these outputs into probabilities via the softmax function to determine the likelihood of each next token. Each logit score for token \( t \) is calculated as:
\begin{equation}
    \left[ \iota _{l,i} \right]_{L \times t}=\begin{bmatrix}
    \iota _{1,i} \\
    \iota _{2,i} \\
    \vdots \\
    \iota _{L,i}
    \end{bmatrix}\times t
\end{equation}
where \( \iota _{l,i} \) denotes the \( l \)-th layer’s logit score for token \( t \) in the generation phase, representing the raw output before applying the softmax.

\subsection{Token Selection}\label{sec:scheme:token_select}

\begin{figure*}[htbp]
    \centering
    \includegraphics[width=\textwidth]{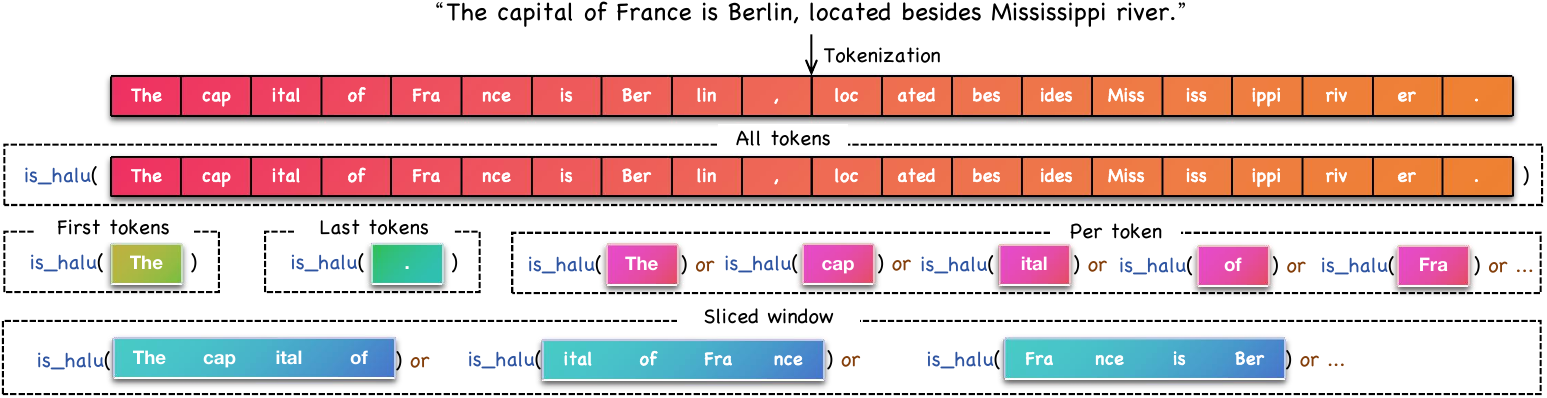}
    \caption{
    Five token selection methods for \sysname:
    (1) All tokens;
    (2) First token;
    (3) Last token;
    (4) Per token;
    (5) Sliced windows.
    }
    \label{fig:token}
\end{figure*}

Unlike previous external methods to detect hallucinations, internal state-based detection relies heavily on selecting tokens.
Each internal state including attention, layer representation, and logit is highly relying on tokens.
Thus we provided five different token selection methods inspired from previous works:

\sssec{All tokens selection}.
Aggregate the features of all tokens (e.g., max, mean) and then input the aggregated result into the $is\_halu$ function to determine if the entire response is hallucinatory.

\[
is\_halu\_all = is\_halu\left(f\left(\{feature(t_i)\}_{i=1}^{N}\right)\right)
\]

where \( t_i \) represents each token in the response, \( N \) is the total number of tokens, and \( f \) is an aggregation function (e.g., max, mean).

\sssec{Per token selection}.
Apply the $is_halu$ function individually to each token. If any token is identified as hallucinatory, the entire response is considered hallucinatory.

\[
is\_halu\_per\_token = \bigvee_{i=1}^{N} is\_halu(t_i)
\]

\sssec{Last token selection}.
Apply the $is_halu$ function only to the last token, using its result to determine if the entire response is hallucinatory.
\[
is\_halu\_last = is\_halu(t_N)
\]
where \( t_N \) is the last token in the response.

\sssec{First token selection}.
Apply the $is_halu$ function only to the first token, using its result to determine if the entire response is hallucinatory.
\[
is\_halu\_first = is\_halu(t_1)
\]
where \( t_1 \) is the first token in the response.

\sssec{Sliced window selection}.
Divide all tokens into multiple sliding windows, where each window contains \( w \) consecutive tokens and slides with a stride \( s \). Apply the $is_halu$ function to each window. If any window is identified as hallucinatory, the entire response is considered hallucinatory.
\[
is\_halu\_sliced = \bigvee_{k=1}^{M} is\_halu(\{t_{k}, t_{k+1}, \dots, t_{k+w-1}\})
\]
where \( M \) is the number of windows, \( w \) is the window size, and \( k \) is the starting index of each window (ranging from 1 to \( N-w+1 \)). The symbol \( \bigvee \) denotes a logical OR operation. If any token group within a window is identified as hallucinatory, the entire response is considered hallucinatory.

We provided examples for each token selection method in Figure \ref{fig:token}.

\subsection{Feature Extraction}\label{sec:scheme:feature_extract}

We extract three groups of features from the extracted layer representations:

\sssec{Attention lookback ratio}.
For each token $t$, we calculate the lookback ratio for each layer and attention head $h$, defined as the proportion of attention score directed to previous tokens among all attention scores for that token:
\[
\frac{\sum_{i < t} \alpha_{l,h,i \rightarrow t}}{\sum_{j} \alpha_{l,h,j \rightarrow t}}
\]
The lookback ratio quantifies the model's focus on historical context. When a model generates hallucinations, it may overlook parts of the contextual information, so a low Lookback Ratio might indicate insufficient use of historical context during generation.

\sssec{Attention allocation sharpness}.
Attention allocation sharpness reflects the concentration of the attention distribution for each token. By computing the entropy of token $t$'s attention distribution in layer $l$ and head $h$, we obtain the Sharpness:
\[
-\sum_{j} p_{l,h,j \rightarrow t} \log p_{l,h,j \rightarrow t}
\]
where
\[
p_{l,h,j \rightarrow t} = \frac{\alpha_{l,h,j \rightarrow t}}{\sum_{k} \alpha_{l,h,k \rightarrow t}}
\]
Attention allocation sharpness indicates whether the model's attention is focused on a few important tokens. Low entropy suggests more concentrated attention, helping us understand if the model might be overly focused on particular tokens, potentially leading to hallucination.

\sssec{Last layer layer representation}.
The last layer representation is extracted directly from the layer representation of the last layer for each token:
\[
\gamma_{L, t}
\]
where $\gamma_{L, t} \in R^d$ represents the layer representation vector of the $L$-th (last) layer for the token $t$. Here:
\begin{itemize}
    \item $d$ is the dimensionality of the model's layer representations (typically the size of the embedding dimension).
    \item $L$ is the total number of transformer layers in the model.
\end{itemize}
The layer representation in the last layer represents the model's final contextual embedding, which directly influences the generated output. Analyzing this layer representation can help identify potential causes of hallucinations in the generated content.

\sssec{Activation map}
The activation map at layer $l$ for a token $t$ is computed from the layer representation $\gamma_{l, t} \in R^d$ as follows:

\begin{enumerate}
    \item Linear Transformation:
    \begin{equation}
    z_{l, t} = W_1 \gamma_{l, t} + b_1,
    \end{equation}
    where $W_1 \in R^{m \times d}$ is the weight matrix, $m$ is the intermediate dimensionality of the feed-forward network, and $b_1 \in R^m$ is the bias vector.

    \item Non-linear Activation:
    \begin{equation}
    a_{l, t} = \text{GELU}(z_{l, t}),
    \end{equation}
    where $a_{l, t} \in R^m$ is the activation map, and $\text{GELU}$ denotes the Gaussian Error Linear Unit activation function.
\end{enumerate}

The resulting activation map $a_{l, t}$ illustrates the degree of activation for $m$ neurons in the feed-forward network at layer $l$ for token $t$.

\sssec{Activation entropy}
Activation entropy is calculated as the entropy of the activation states in each layer, capturing the distribution of activations:
\[
-\sum_{j} p_{l,j \rightarrow t} \log p_{l,j \rightarrow t}
\]
where
\[
p_{l,j \rightarrow t} = \frac{a_{l,j \rightarrow t}}{\sum_{k} a_{l,k \rightarrow t}}
\]
Activation entropy reflects the dispersion or concentration of the model's activation states. Low entropy indicates focused activation on a few tokens, which could signify overconfidence in certain tokens, potentially leading to hallucination.

\sssec{Min token probs}
Min token probs represent the lowest softmax probability across tokens in each layer's logit output:
\[
\min_{t} \left( \text{softmax}(\iota_{l, t}) \right)
\]
Low probability values indicate a lack of confidence in certain tokens. These extremely low probabilities can help identify potential points of hallucination risk.

\sssec{Max token ranks}
Max token ranks represent the rank of the token with the lowest probability across a sequence within each layer’s softmax logits:
\[
\max_{t} \left( \text{rank}(\text{softmax}(\iota_{l, t})) \right)
\]
The token with the lowest rank may indicate low-priority content during generation. High-rank values suggest that the model may hallucinate by producing low-priority tokens in the sequence.

\sssec{Joint token probs}.
Joint token probs represent the product of probabilities of all tokens within a sequence in each layer’s softmax logits:
\[
\prod_{t} \text{softmax}(\iota_{l, t})
\]
Joint token probs provides a measure of overall confidence in generating the sequence. A low joint probability suggests low model confidence in the entire sequence, which could result in hallucinated content.

\section{Understanding Internal State}\label{sec:understand}

In this section, we provide an understanding of how the internal state changes during the lens of LLM inference (\S\ref{sec:understand:infer}), inference with RAG (\S\ref{sec:understand:rag}).

\subsection{Understanding through Lens of Inference}\label{sec:understand:infer}

\begin{table*}[h!]
\centering
\begin{tabular}{|M{3cm}|M{3cm}|M{3cm}|M{3cm}|M{3cm}|}
\hline\hline
\multicolumn{5}{|c|}{
\includegraphics[width=9cm]{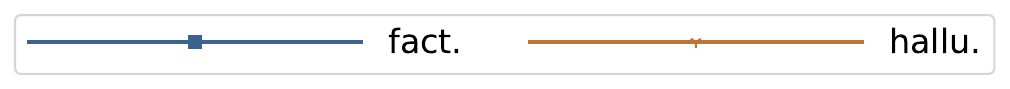}
}
\\
\hline
\multicolumn{2}{|c|}{Attention} & \multicolumn{1}{|c|}{Activation} & \multicolumn{2}{|c|}{Logit} \\
\hline\hline

\multicolumn{5}{|c|}{Dataset: HaluEval} \\
\hline

Lookback ratio & Attention entropy & Hidden states & Min. token prob. & Joint token prob. \\
\hline

\includegraphics[width=3cm]{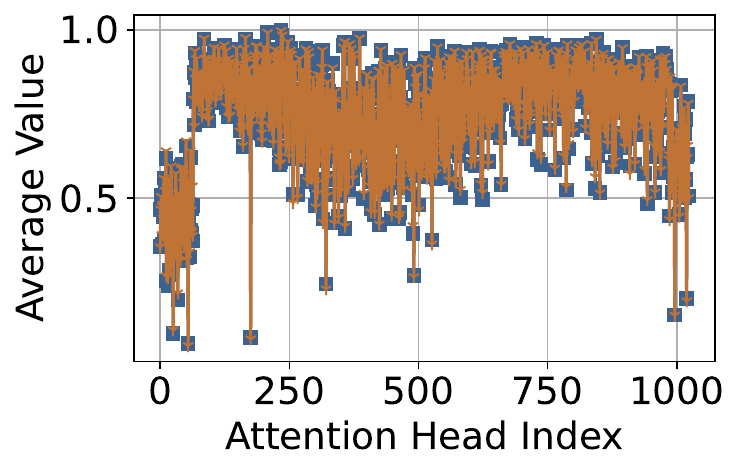} &
\includegraphics[width=3cm]{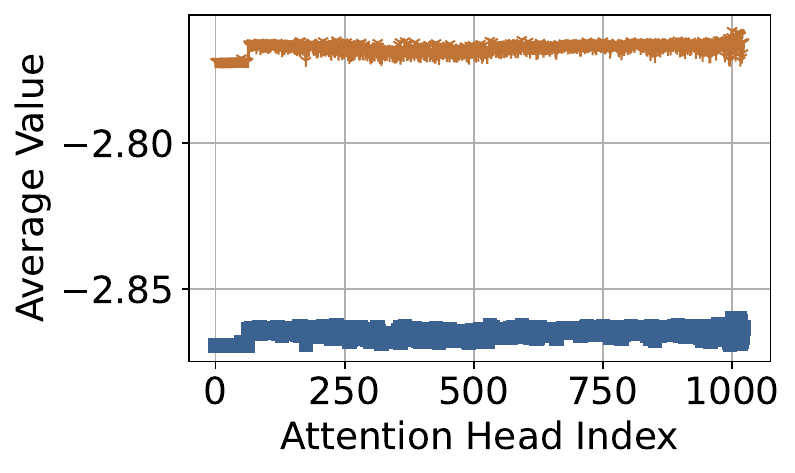} &
\includegraphics[width=3cm]{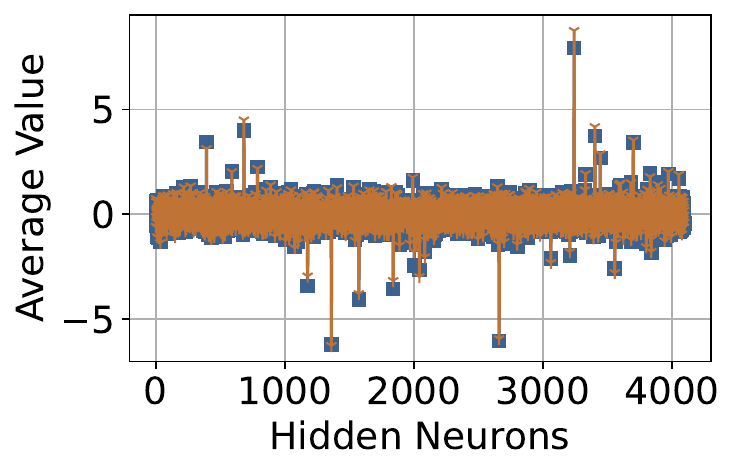} &
\includegraphics[width=3cm]{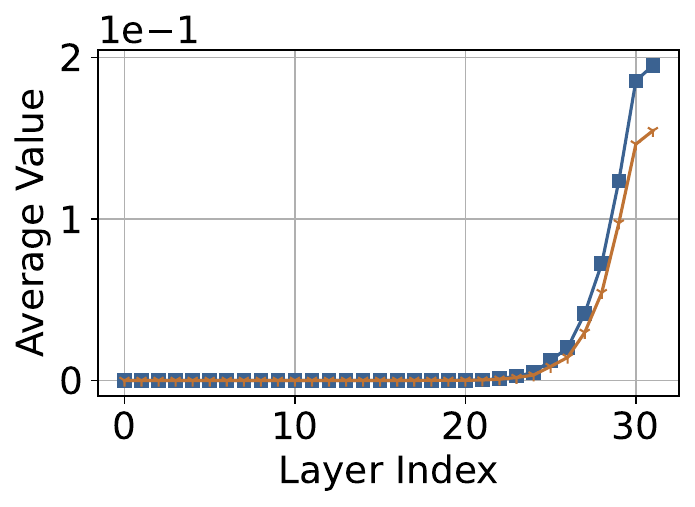} &
\includegraphics[width=3cm]{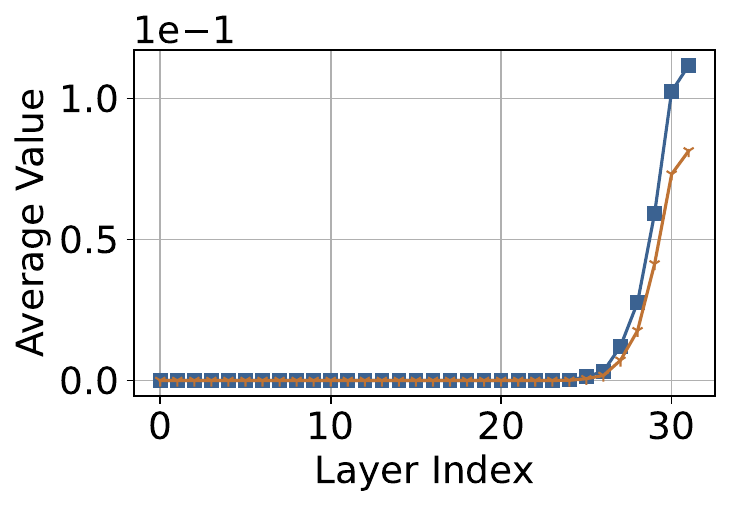} \\
\hline

Lookback ratio & Attention entropy & Activation sharpness & Max. token rank & Avg. dist. divergence \\
\hline

\includegraphics[width=3cm]{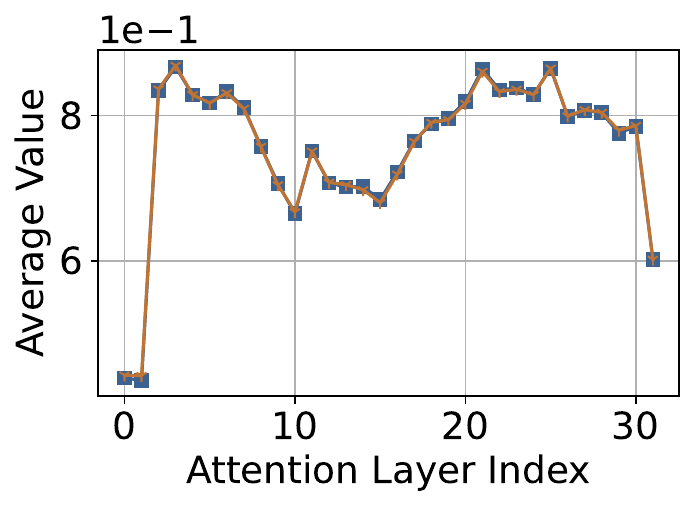} &
\includegraphics[width=3cm]{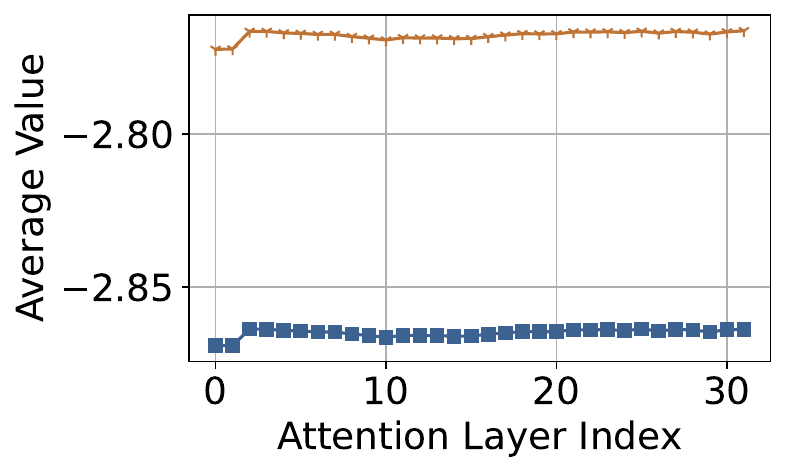} &
\includegraphics[width=3cm]{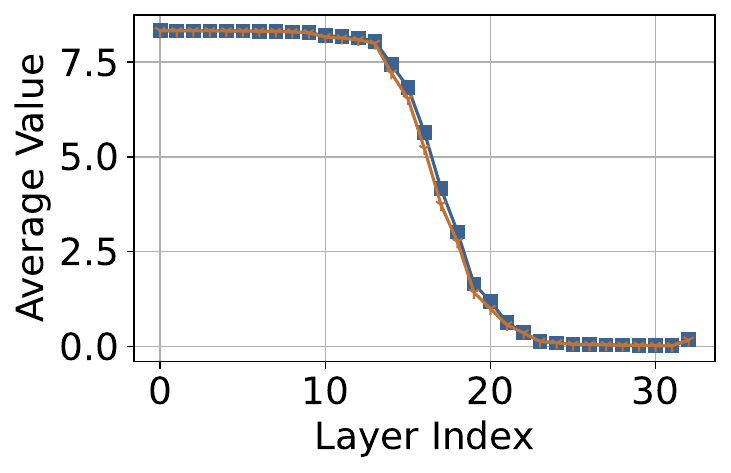} &
\includegraphics[width=3cm]{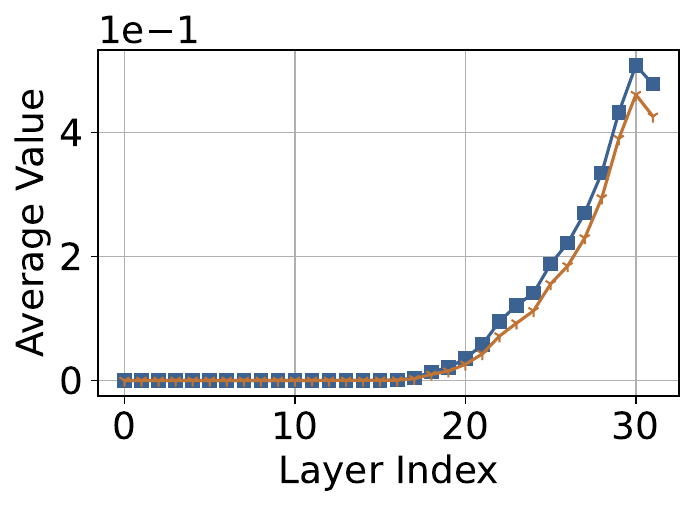} &
\includegraphics[width=3cm]{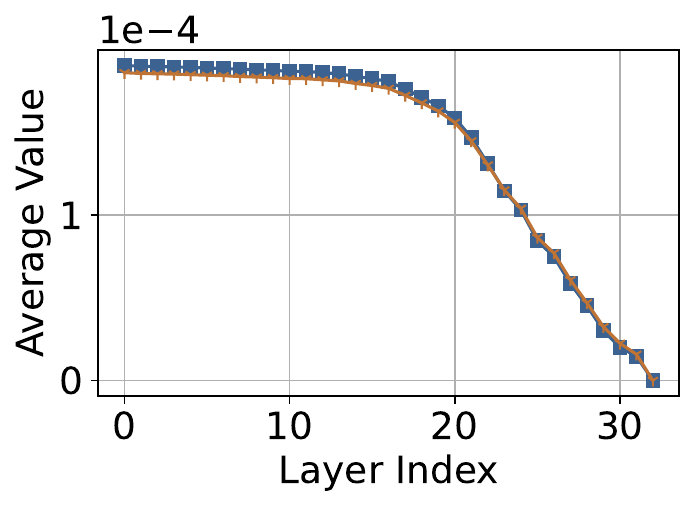} \\

\hline\hline

\multicolumn{5}{|c|}{Dataset: CNNDM} \\
\hline

Lookback ratio & Attention entropy & Hidden states & Min. token prob. & Joint token prob. \\
\hline

\includegraphics[width=3cm]{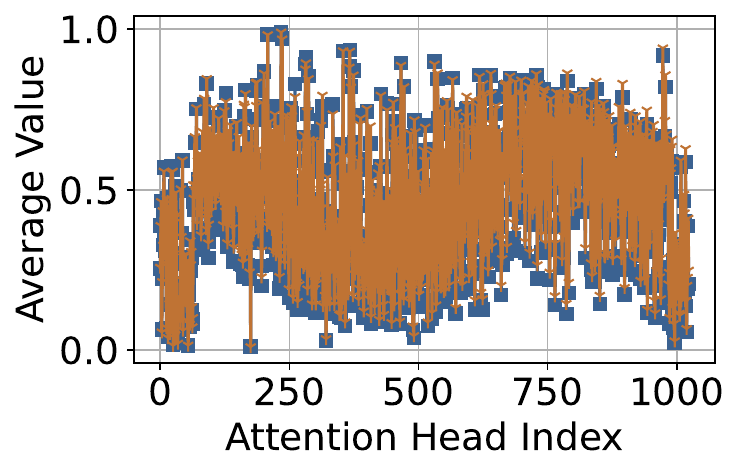} &
\includegraphics[width=3cm]{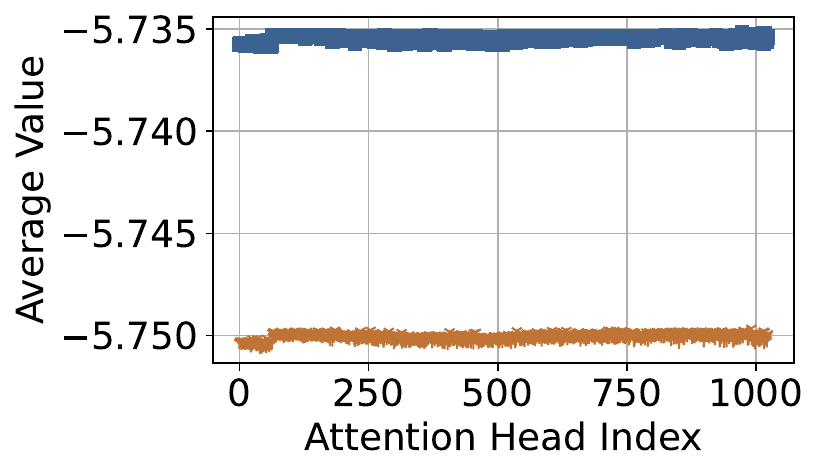} &
\includegraphics[width=3cm]{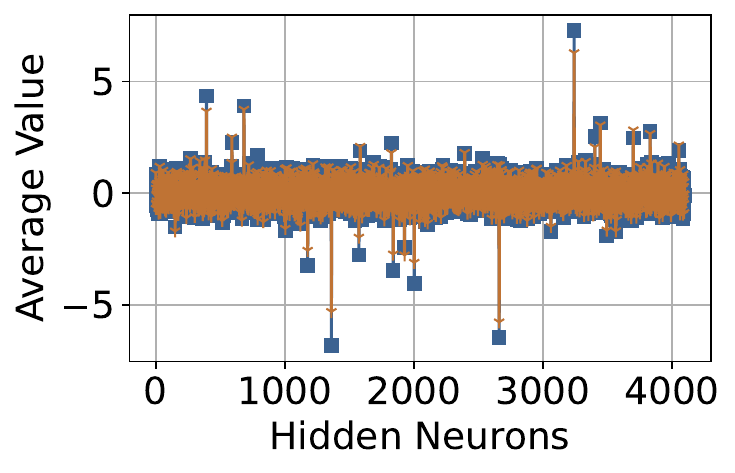} &
\includegraphics[width=3cm]{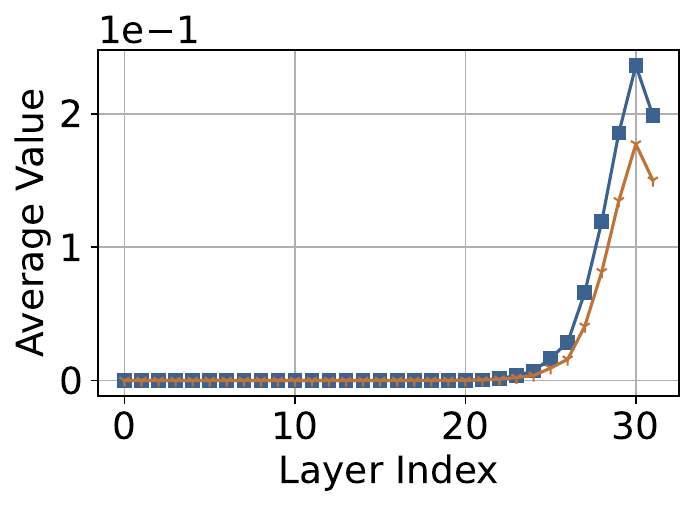} &
\includegraphics[width=3cm]{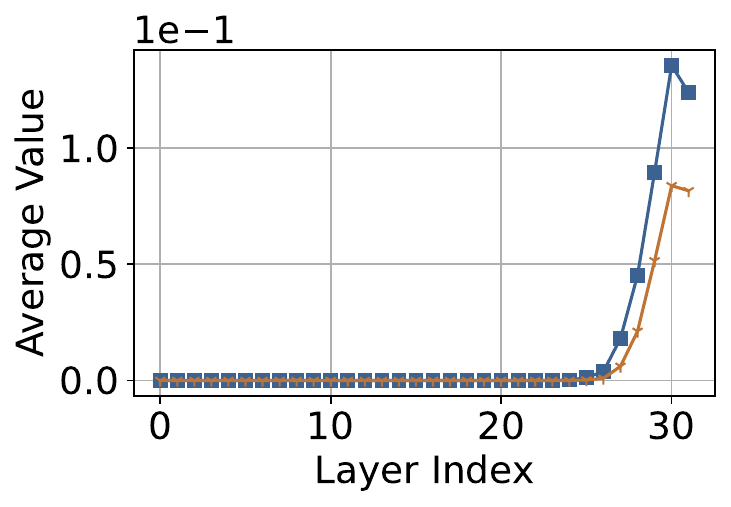} \\
\hline

Lookback ratio & Attention entropy & Activation sharpness & Max. token rank & Avg. dist. divergence \\
\hline

\includegraphics[width=3cm]{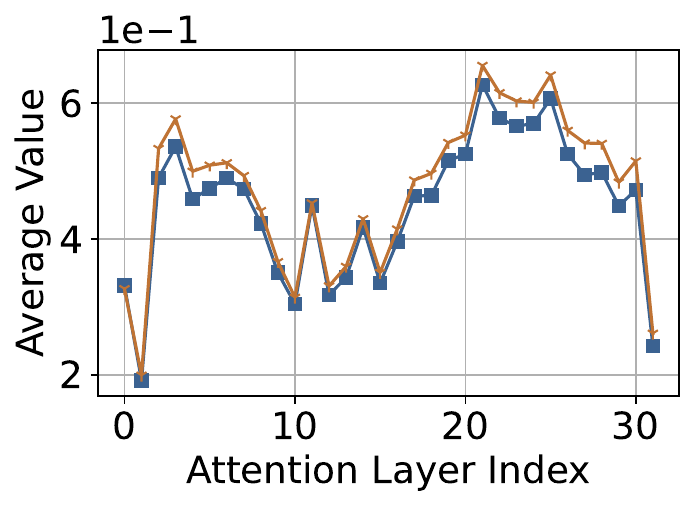} &
\includegraphics[width=3cm]{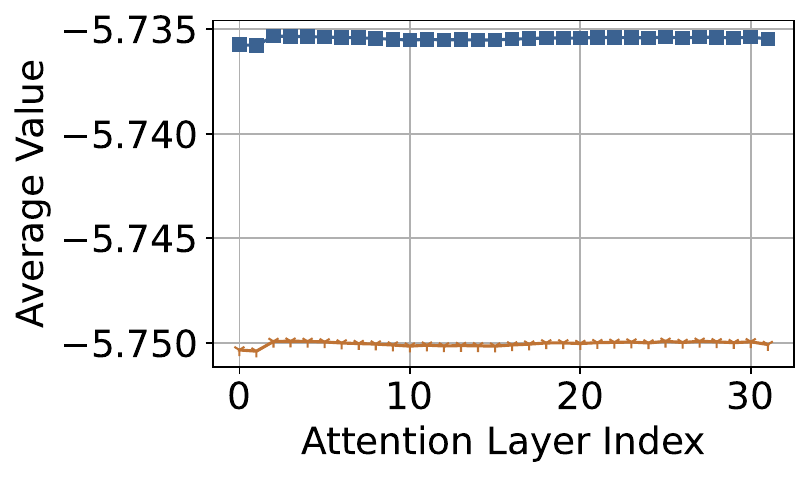} &
\includegraphics[width=3cm]{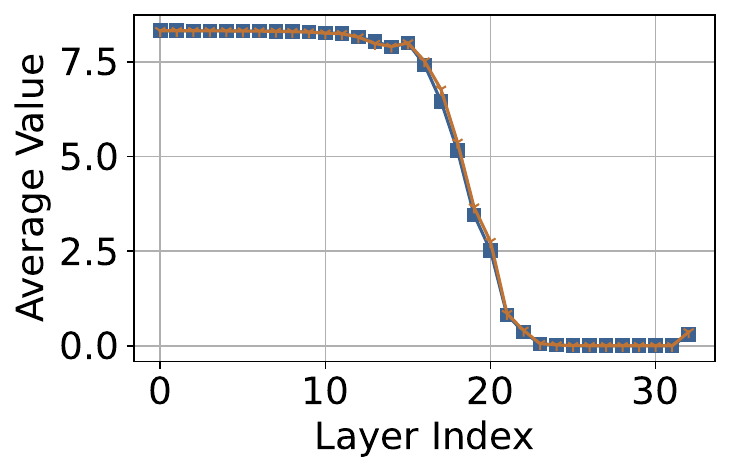} &
\includegraphics[width=3cm]{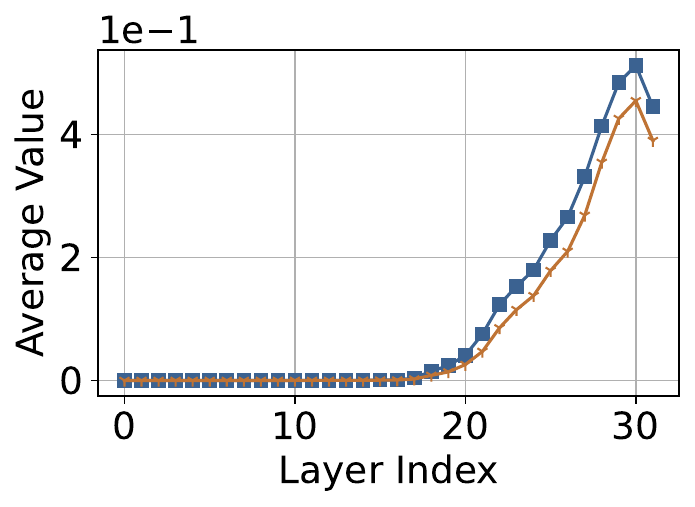} &
\includegraphics[width=3cm]{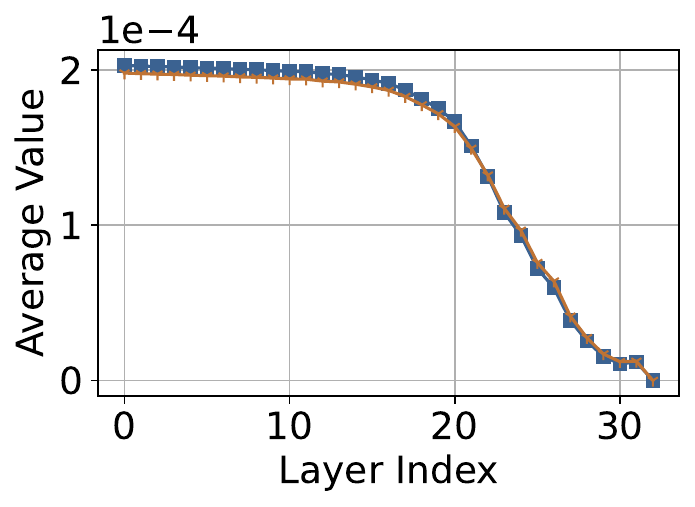} \\

\hline\hline

\end{tabular}
\caption{We visualize the comparison of attention-type features between inference with RAG and without RAG.}
\label{fig:understand:infer}
%\vspace{-15pt}
\end{table*}

\begin{figure}[t]
\centering
      \begin{minipage}{\columnwidth}
          {\centering{\hspace{1.5cm}CNNDM\hspace{3cm}HaluEval}}
      \end{minipage}
      \subfloat[Avg. jsd]{\includegraphics[width=0.5\columnwidth]{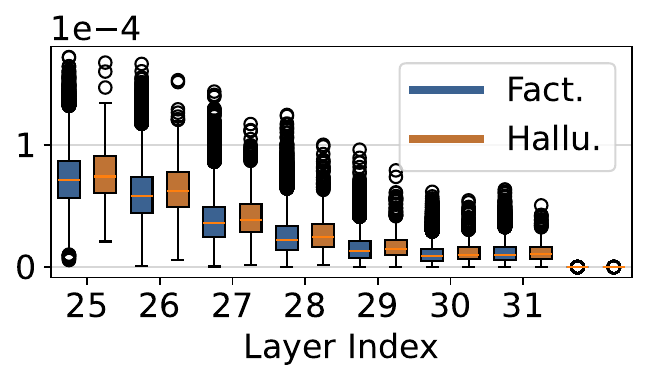}}
      \subfloat[Avg. jsd]{\includegraphics[width=0.5\columnwidth]{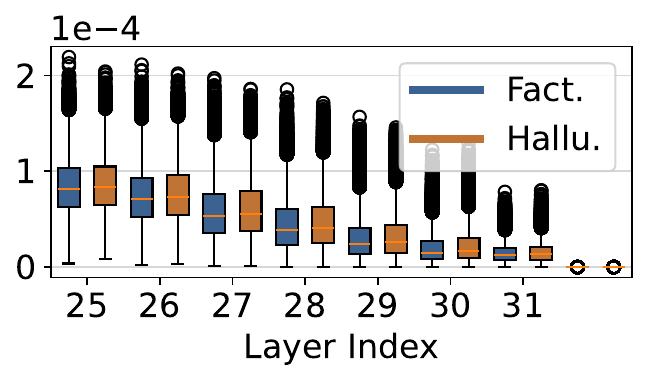}}
      \\
      \subfloat[Min. prob.]{\includegraphics[width=0.5\columnwidth]{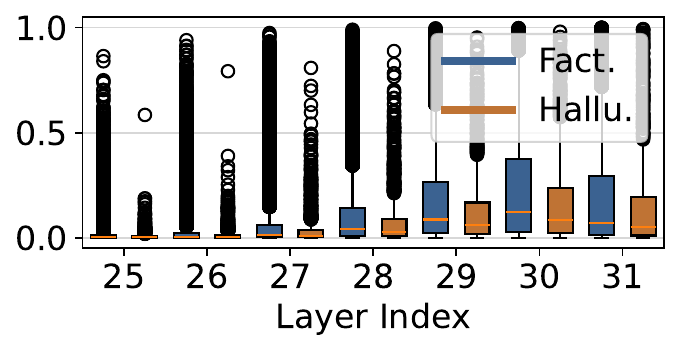}}
      \subfloat[Min. prob.]{\includegraphics[width=0.5\columnwidth]{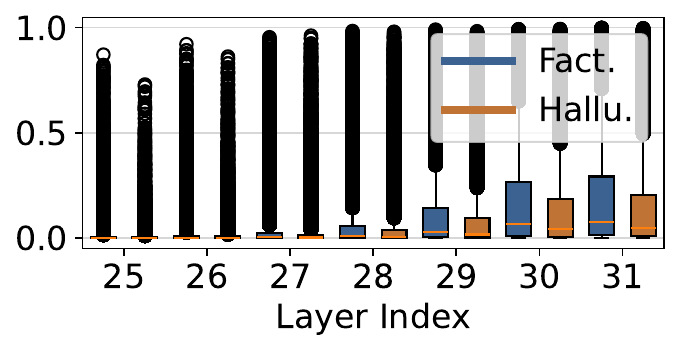}}
      \\
      \subfloat[Joint prob.]{\includegraphics[width=0.5\columnwidth]{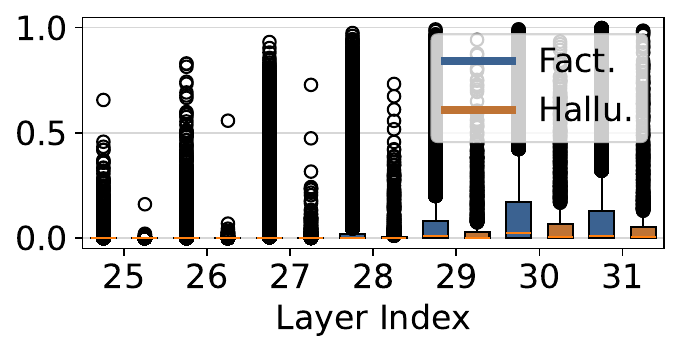}}
      \subfloat[Joint prob.]{\includegraphics[width=0.5\columnwidth]{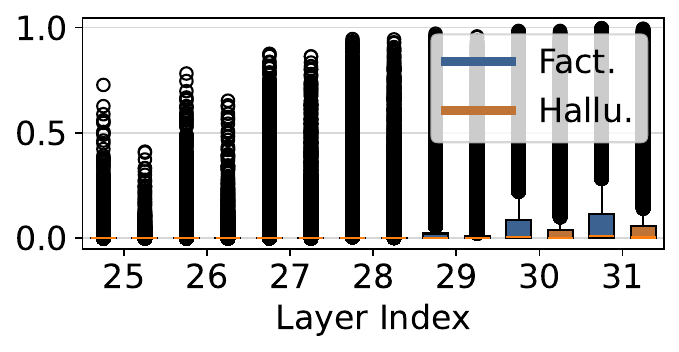}}
      \\
      \subfloat[Max. rank]{\includegraphics[width=0.5\columnwidth]{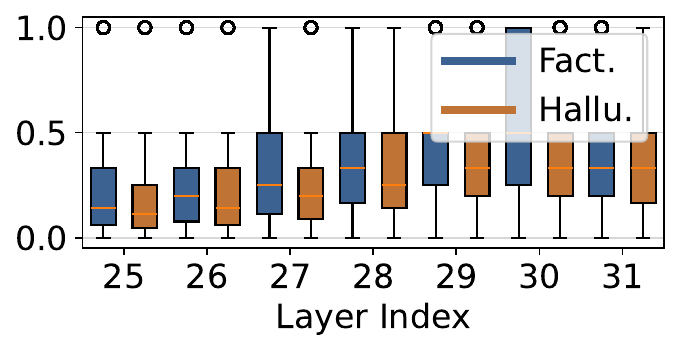}}
      \subfloat[Max. rank]{\includegraphics[width=0.5\columnwidth]{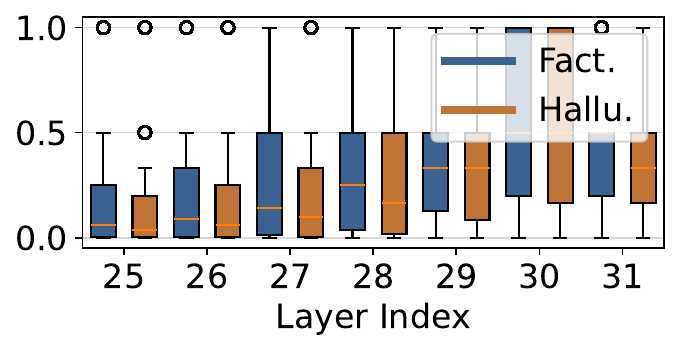}}
      \\
      \caption{
      For the logit class internal state features, some layers of features seem to be very discriminative. We visualize the specific distribution of all samples of the features of these layers and show them in the figure. It can be seen that although there is a big difference in the overall mean, the overall distribution of this type of features of the hallucination output and the actual output overlap, and it is difficult to say that it is a very discriminative feature.
      }
    \label{fig:understand:key:logit}
\end{figure}

\sssec{Method}.
We experimented to compare internal states that differ from the factual output and hallucinated output across the whole inference process.
Specifically, we used 2 datasets: HaluEval for factual hallucination generation, and CNNDM for faithful hallucination generation.
We collected the internal states of tested models using the 2 datasets and computed the extracted features.
At last, we compared internal states between hallucinated output and factual output.
The results are shown in Table \ref{fig:understand:infer}.

\sssec{Results on attention states}.  
The lookback ratio (\(\text{LR}\)) measures the proportion of attention directed toward previous tokens. In the HaluEval dataset, both factual (\(\mathcal{F}\)) and hallucinated (\(\mathcal{H}\)) outputs exhibit overlapping trends across attention heads (\(h \in H\)) and attention layers (\(l \in L\)), implying that \(\text{LR}_{l,h}\) does not effectively distinguish between \(\mathcal{F}\) and \(\mathcal{H}\). However, in the CNNDM dataset, \(\text{LR}_{l,h}\) demonstrates a clearer separation, with \(\mathcal{F}\) maintaining higher and more stable values compared to \(\mathcal{H}\) (\( \text{LR}^{\mathcal{F}}_{l,h} > \text{LR}^{\mathcal{H}}_{l,h}, \forall l,h\)). This suggests that hallucinations in CNNDM are associated with reduced attention to prior context, making \(\text{LR}\) a more sensitive feature for hallucination detection in this dataset.

The attention entropy (\(\mathcal{E}_l\)) quantifies the concentration of attention distributions. For HaluEval, results show that \(\mathcal{F}\) exhibits lower entropy (\(\mathcal{E}_l^{\mathcal{F}} < \mathcal{E}_l^{\mathcal{H}}\)), indicating more focused attention, whereas \(\mathcal{H}\) outputs have higher entropy, reflecting dispersed attention. Aggregated across layers (\(\bar{\mathcal{E}}_l\)), this trend persists. In CNNDM, although \(\mathcal{E}_l^{\mathcal{H}} > \mathcal{E}_l^{\mathcal{F}}\), the distinction is less pronounced. These results suggest that \(\mathcal{E}_l\) robustly captures the dispersion of attention in \(\mathcal{H}\), particularly for HaluEval.

In summary, \(\text{LR}\) is dataset-sensitive, effectively distinguishing hallucinations in CNNDM but not in HaluEval, whereas \(\mathcal{E}_l\) consistently differentiates between \(\mathcal{H}\) and \(\mathcal{F}\) in both datasets, highlighting dispersed attention as a hallmark of hallucinations.

\sssec{Results on activation states}.  
The hidden states (\(h_{l,t}\)) represent activation values at layer \(l\) for token \(t\). Across both datasets, factual and hallucinated outputs exhibit overlapping trends in \(\bar{h}_{l,t}\), indicating that \(\bar{h}_{l,t}^{\mathcal{F}} \approx \bar{h}_{l,t}^{\mathcal{H}}\). This overlap suggests that \(h_{l,t}\) lacks sensitivity to the nuanced differences introduced by hallucinations.

Activation sharpness (\(S_l\)) measures the concentration of activations across layers. In both datasets, \(S_l^{\mathcal{F}}\) and \(S_l^{\mathcal{H}}\) follow nearly identical trends, with a consistent decrease after the 20th layer (\(\forall l > 20, S_l^{\mathcal{F}} \approx S_l^{\mathcal{H}}\)). This similarity suggests that activation sharpness fails to capture hallucination-specific patterns. The uniformity in \(\bar{S}_l\) across \(\mathcal{F}\) and \(\mathcal{H}\) implies that these features reflect general activation dynamics rather than hallucination-induced variations.

Thus, neither \(h_{l,t}\) nor \(S_l\) effectively distinguishes between hallucinations and factual outputs, underscoring their limited utility for this task.

\begin{figure}t]
\centering
      \subfloat[CNNDM]{\includegraphics[width=0.5\columnwidth]{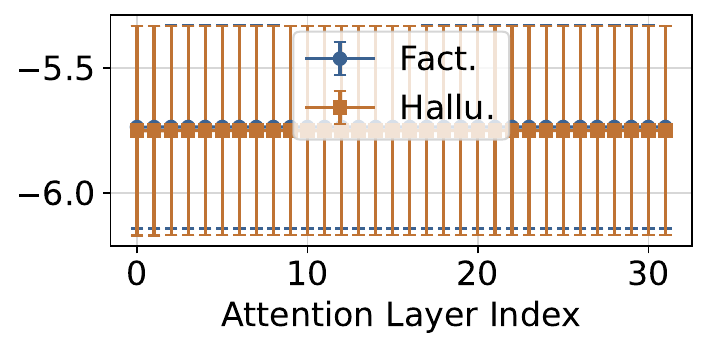}}
      \subfloat[HaluEval]{\includegraphics[width=0.5\columnwidth]{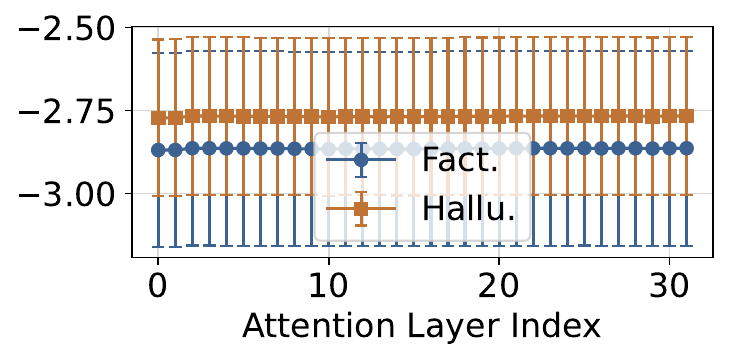}}
      \caption{
      For the attention entropy internal state features, we visualize the specific distribution of all samples of the features of these layers and show them in the figure.
      }
    \label{fig:understand:key:atten_entropy}
\end{figure}

\sssec{Results on logit states}.  
The minimum token probability (\(\min(\text{P}_{l,t})\)) captures the model's confidence in its least likely predicted token. In HaluEval, \(\min(\text{P}_{l,t}^{\mathcal{H}}) < \min(\text{P}_{l,t}^{\mathcal{F}})\), but the difference is slight, whereas in CNNDM, this disparity is more pronounced (\(\forall l, \min(\text{P}_{l,t}^{\mathcal{H}}) \ll \min(\text{P}_{l,t}^{\mathcal{F}})\)), indicating greater uncertainty in hallucinated outputs.

The joint token probability (\(\prod_t \text{P}_{l,t}\)) reflects cumulative confidence. In HaluEval, \(\prod_t \text{P}_{l,t}^{\mathcal{F}} \approx \prod_t \text{P}_{l,t}^{\mathcal{H}}\), while in CNNDM, hallucinated outputs exhibit lower joint probabilities (\(\prod_t \text{P}_{l,t}^{\mathcal{H}} < \prod_t \text{P}_{l,t}^{\mathcal{F}}\)) in later layers.

The maximum token rank (\(\max(\text{R}_{l,t})\)) indicates the rank of the least confident token. In HaluEval, \(\max(\text{R}_{l,t})\) trends for \(\mathcal{F}\) and \(\mathcal{H}\) largely overlap (\(\max(\text{R}_{l,t}^{\mathcal{F}} \approx \max(\text{R}_{l,t}^{\mathcal{H}})\)), but in CNNDM, hallucinated outputs consistently exhibit higher ranks (\(\max(\text{R}_{l,t}^{\mathcal{H}}) > \max(\text{R}_{l,t}^{\mathcal{F}}\)).

The Jensen-Shannon Divergence (JSD) measures distribution divergence. In both datasets, JSD trends for \(\mathcal{F}\) and \(\mathcal{H}\) are nearly identical (\(\text{JSD}_{l,t}^{\mathcal{F}} \approx \text{JSD}_{l,t}^{\mathcal{H}}\)), indicating limited sensitivity to hallucination-specific variations.

In conclusion, logit features such as \(\min(\text{P}_{l,t})\) and \(\max(\text{R}_{l,t})\) are more effective in CNNDM than in HaluEval, reflecting the stronger correlation between reduced confidence and hallucinations in CNNDM. However, features like \(\prod_t \text{P}_{l,t}\) and JSD show limited efficacy across both datasets.

\subsection{Impact of RAG during Inference}\label{sec:understand:rag}

\begin{table*}[h!]
\centering
\begin{tabular}{|M{3cm}|M{3cm}|M{3cm}|M{3cm}|M{3cm}|}
\hline\hline
\multicolumn{5}{|c|}{
\includegraphics[width=9cm]{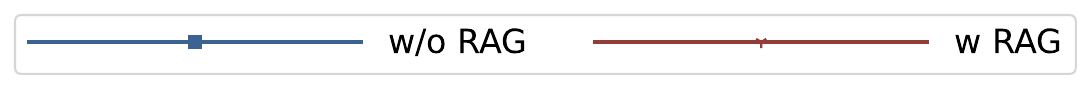}
}
\\
\hline
\multicolumn{2}{|c|}{Attention} & \multicolumn{1}{|c|}{Activation} & \multicolumn{2}{|c|}{Logit} \\
\hline\hline

Lookback ratio & Attention entropy & Hidden states & Min. token prob. & Joint token prob. \\
\hline

\includegraphics[width=3cm]{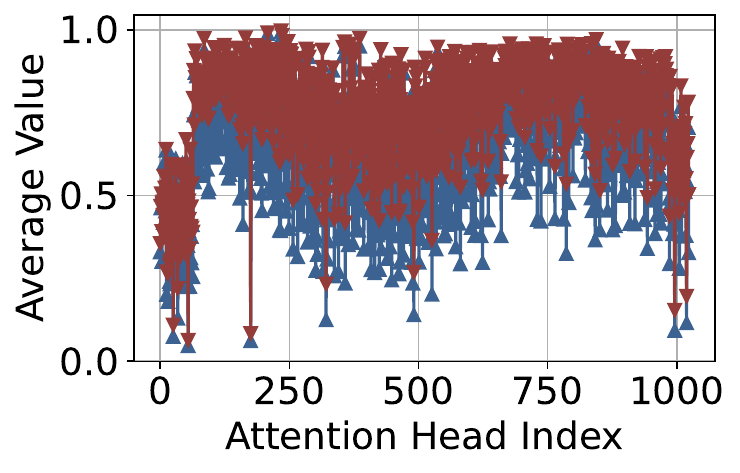} &
\includegraphics[width=3cm]{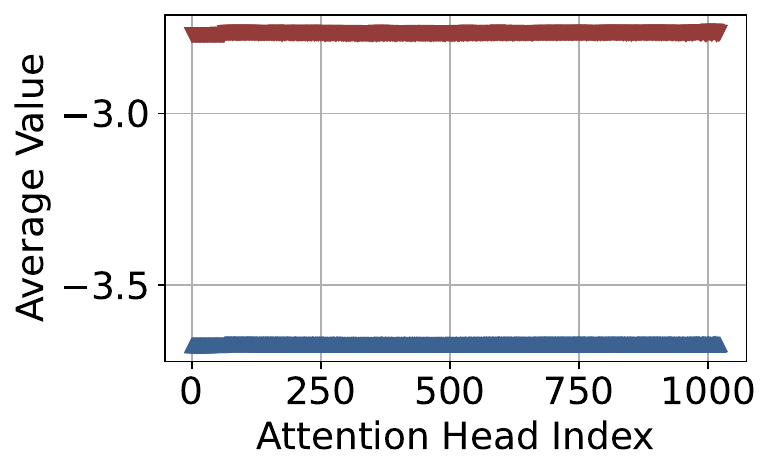} &
\includegraphics[width=3cm]{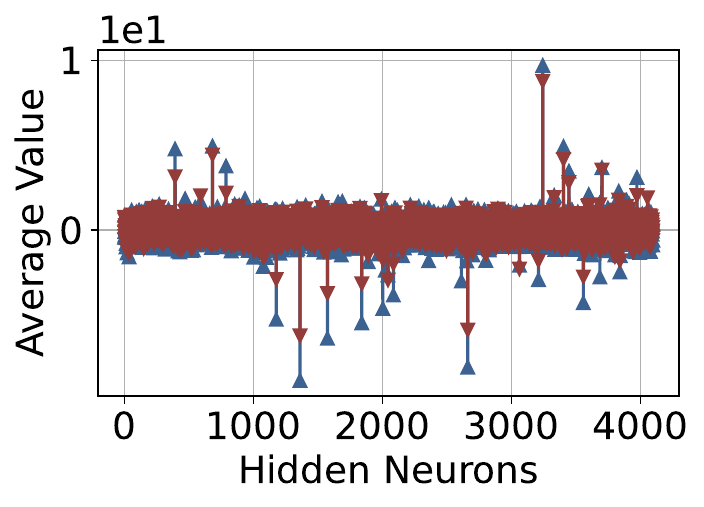} &
\includegraphics[width=3cm]{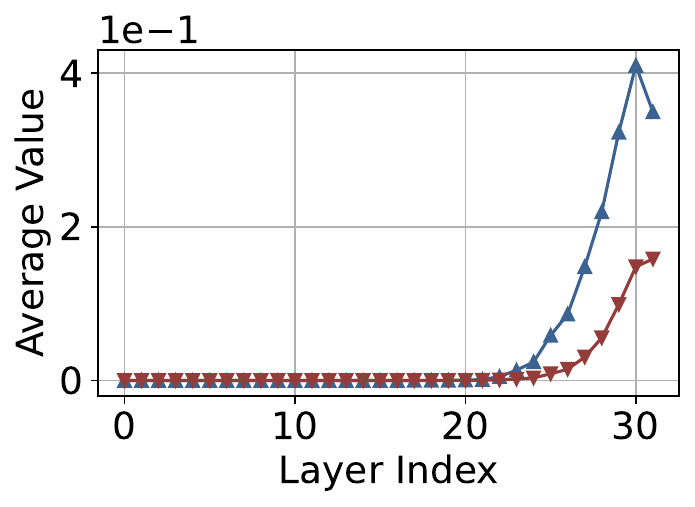} &
\includegraphics[width=3cm]{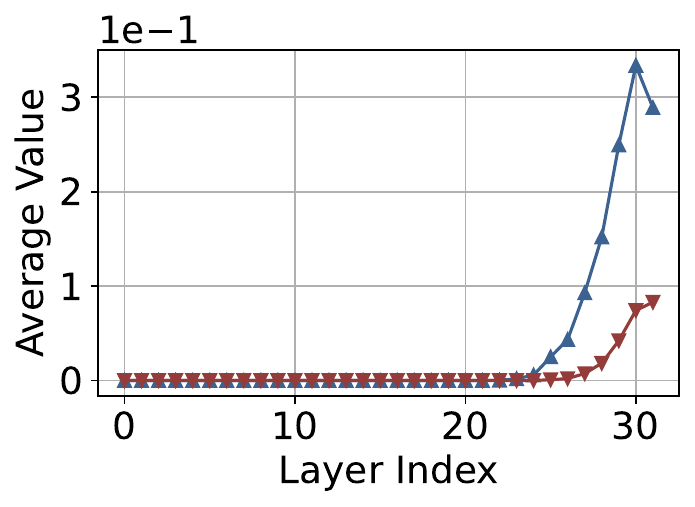} \\
\hline

Lookback ratio & Attention entropy & Activation sharpness & Max. token rank & Avg. dist. divergence \\
\hline

\includegraphics[width=3cm]{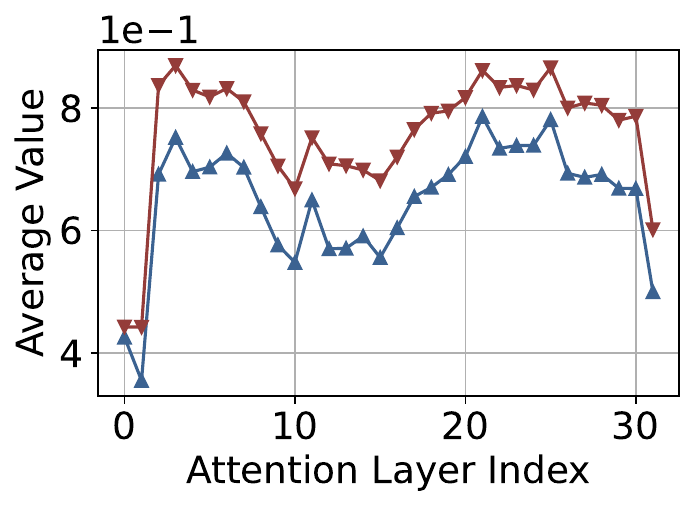} &
\includegraphics[width=3cm]{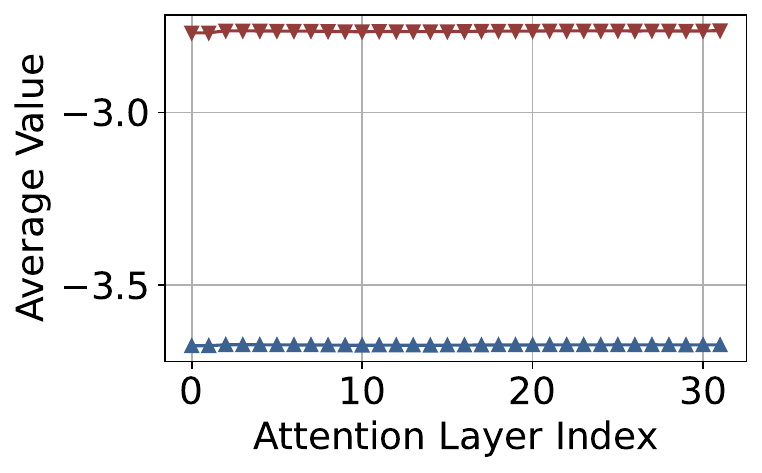} &
\includegraphics[width=3cm]{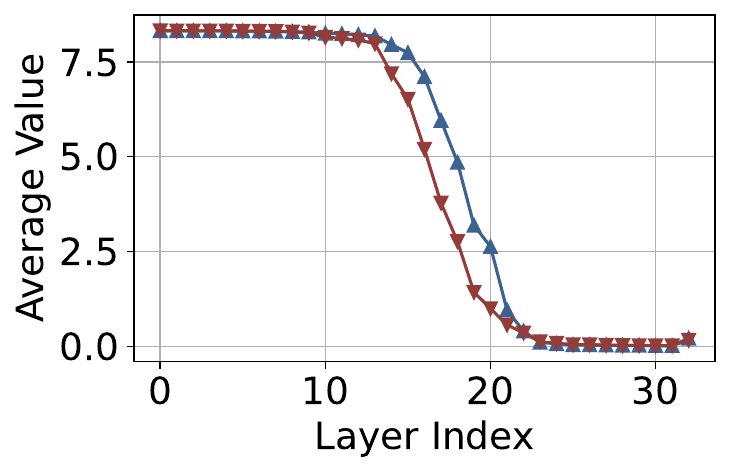} &
\includegraphics[width=3cm]{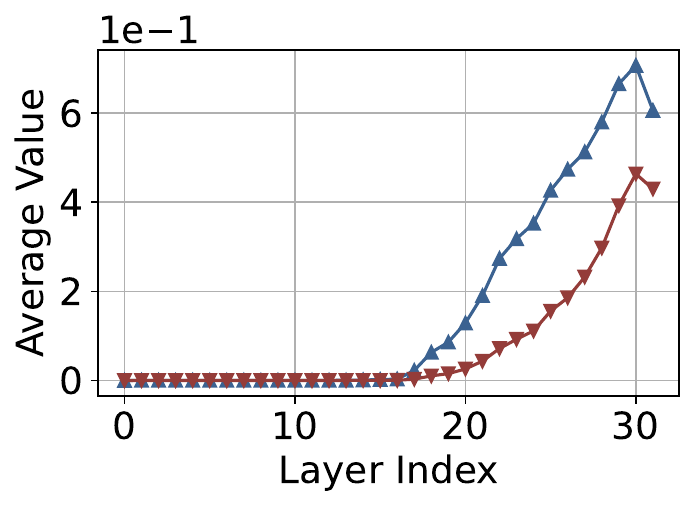} &
\includegraphics[width=3cm]{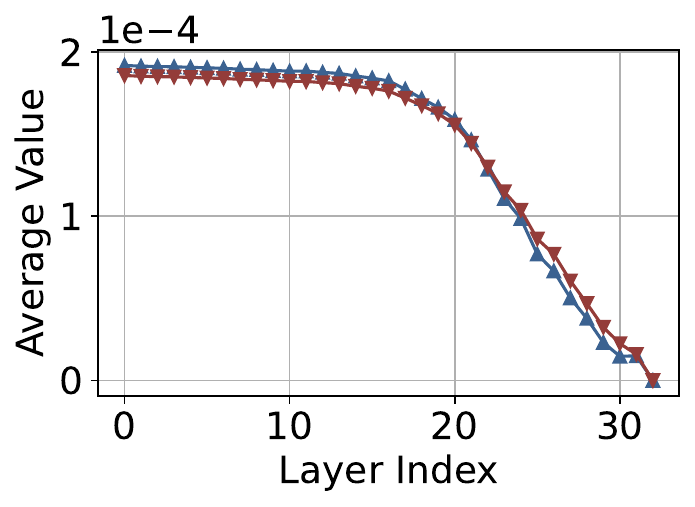} \\

\hline\hline

\end{tabular}
\caption{We visualize the comparison of attention-type features between inference with RAG and without RAG.}
\label{fig:understand:rag}
%\vspace{-15pt}
\end{table*}

\sssec{Method}.
We utilized the HaluEval dataset to investigate the effect of RAG on LLM inference. For each question, the answer was converted into a retrieval-augmented knowledge base (RAG) to assist the LLM. Internal states, including attention scores, hidden states, and logits, were extracted during inference under two settings: with and without RAG. Features such as attention lookback ratio, activation entropy, and token probabilities were computed and compared between the two conditions, focusing on correct responses to analyze how RAG influences the model’s reasoning process and mitigates hallucinations.

\sssec{Results on attention states}.  
The lookback ratio (\(\text{LR}_{l,h}\)) reflects the model's focus on prior tokens. At the head level, the figure shows the distribution of \(\text{LR}_{l,h}\) across all 1024 attention heads (\(h \in H\)), while at the layer level, it aggregates over the 32 heads per layer (\(l \in L\)). The results indicate that \(\bar{\text{LR}}_l^{\text{RAG}}\) (with RAG) is more consistent and slightly higher than \(\bar{\text{LR}}_l^{\text{non-RAG}}\) (without RAG), particularly across deeper layers (\(l > 10\)). This consistency arises because RAG provides external context, enhancing backward focus and enabling effective utilization of retrieved information. Without RAG, the model depends entirely on internal context, leading to greater variability (\(\text{Var}(\bar{\text{LR}}_l^{\text{non-RAG}}) > \text{Var}(\bar{\text{LR}}_l^{\text{RAG}})\)).

The attention entropy (\(\mathcal{E}_{l,h}\)) quantifies the diversity of attention distributions. At the head level, \(\mathcal{E}_{l,h}^{\text{RAG}}\) is consistently higher than \(\mathcal{E}_{l,h}^{\text{non-RAG}}\), and at the layer level, the average entropy \(\bar{\mathcal{E}}_l^{\text{RAG}}\) also demonstrates broader focus compared to \(\bar{\mathcal{E}}_l^{\text{non-RAG}}\). This is because RAG enriches attention mechanisms with external knowledge, allowing attention to be allocated across a wider set of tokens (\(\mathcal{E}_{l,h}^{\text{RAG}} > \mathcal{E}_{l,h}^{\text{non-RAG}}, \forall l,h\)). Conversely, without RAG, attention is constrained to narrower contexts, resulting in lower entropy values.

\sssec{Results on activation states}.  
The hidden states (\(h_{l,t}\)) represent the activation magnitudes across neurons for token \(t\) at layer \(l\). At the neuron level, the distribution of \(\text{Avg}(h_{l,t})\) shows that RAG stabilizes activations, reducing variability (\(\text{Var}(h_{l,t}^{\text{RAG}}) < \text{Var}(h_{l,t}^{\text{non-RAG}})\)). At the layer level, the average activations (\(\bar{h}_l\)) are consistently higher with RAG (\(\bar{h}_l^{\text{RAG}} > \bar{h}_l^{\text{non-RAG}}\)) due to the additional context provided by RAG, which enhances the magnitude and consistency of neuron activations. Without RAG, activations are unstable, occasionally exhibiting outliers.

Activation sharpness (\(S_l\)), which reflects the concentration of activation values, is higher in earlier layers (\(l \leq 10\)) for both RAG and non-RAG settings. However, the decline in sharpness is more gradual with RAG, maintaining sharper activations (\(S_l^{\text{RAG}} > S_l^{\text{non-RAG}}, \forall l > 10\)). This phenomenon suggests that RAG helps preserve activation focus across deeper layers, likely due to the inclusion of external context. Without RAG, sharpness deteriorates more rapidly as activations diffuse across the network.

\sssec{Results on logit states}.  
The minimum token probability (\(\min(\text{P}_{l,t})\)) represents the model's confidence in its least likely predicted token. Both the head-level and layer-level trends show that \(\min(\text{P}_{l,t})\) increases sharply in later layers (\(l > 10\)). The difference between RAG and non-RAG settings is negligible (\(\min(\text{P}_{l,t}^{\text{RAG}} \approx \min(\text{P}_{l,t}^{\text{non-RAG}})\)), indicating that this feature is primarily influenced by internal computations rather than external context.

The joint token probability (\(\prod_t \text{P}_{l,t}\)) reflects overall confidence in generating a sequence. Across layers, \(\prod_t \text{P}_{l,t}^{\text{RAG}} > \prod_t \text{P}_{l,t}^{\text{non-RAG}}\), with the gap being more pronounced in later layers (\(l > 15\)). This improvement arises from RAG's ability to provide richer context, enhancing the model's cumulative confidence in token generation.

The average distribution divergence (\(\text{D}_{\text{avg}}\)) measures uncertainty by comparing predicted and reference distributions. In both RAG and non-RAG settings, \(\text{D}_{\text{avg}}\) decreases consistently across layers (\(l > 5\)), showing that uncertainty is reduced through layer-wise refinement. The difference between RAG and non-RAG is minimal (\(\text{D}_{\text{avg}}^{\text{RAG}} \approx \text{D}_{\text{avg}}^{\text{non-RAG}}\)), suggesting that uncertainty reduction is primarily driven by internal processes.

The maximum token rank (\(\max(\text{R}_{l,t})\)) captures the relative position of the least confident token in the ranking. Across layers, \(\max(\text{R}_{l,t})\) decreases consistently, reflecting growing confidence in top predictions. There is no significant distinction between RAG and non-RAG (\(\max(\text{R}_{l,t}^{\text{RAG}} \approx \max(\text{R}_{l,t}^{\text{non-RAG}})\)), indicating that token ranking is predominantly determined by internal model dynamics.

%\subsection{Impact of Fine-tuning during Inference}\label{sec:understand:finetune}

%\sssec{Method}.

%\subsection{Feature Correlation Analysis}\label{sec:understand:corr}
\section{\sysname's Performance on Detection}\label{sec:detect}

\subsection{Datasets}\label{sec:detect:datasets}

\begin{table*}[]
\normalsize
\begin{tabular}{|M{1.5cm}|M{1.2cm}|M{3.2cm}|M{1.1cm}|M{1.4cm}|M{1.4cm}|M{1.1cm}|M{1.4cm}|M{1.4cm}|}
\hline\hline
\multirow{2}{*}{Models} & \multirow{2}{*}{Groups} & \multirow{2}{*}{Features} & \multicolumn{3}{c|}{HaluEval}                    & \multicolumn{3}{c|}{CNNDM}                       \\ \cline{4-9}
& & & Accuracy & Recall(Halu) & Recall(Fact) & Accuracy & Recall(Halu) & Recall(Fact) \\
\hline\hline
\multirow{10}{*}{Llama-2-7B} & \multirow{4}{*}{Logit}      & Max token rank            & 0.53 & 0.31 & 0.74 & 0.60 & 0.60 & 0.60 \\
& & Min token prob            & 0.54 & 0.64 & 0.44 & 0.58 & 0.63 & 0.57 \\
& & Joint prob                & 0.54 & 0.72 & 0.37 & 0.63 & 0.50 & 0.67 \\
& & JSD                       & 0.51 & 0.00 & 1.00 & 0.51    & 0.49    & 0.52    \\

\cline{2-9}\cline{2-9}
& \multirow{3}{*}{Attention}  & Lookback ratio            & \textbf{0.75} & 0.75 & 0.74 & \textbf{0.83} & 0.77 & 0.84 \\
& & Attention sharpness       & 0.60 & 0.58 & 0.63 & 0.61 & 0.54 & 0.63 \\
& & Key token attention ratio & 0.70 & 0.68 & 0.72 & 0.72    & 0.72    & 0.71    \\

\cline{2-9}\cline{2-9}
& \multirow{3}{*}{Activation} & Hidden state              & \textbf{0.76} & 0.76 & 0.76 & \textbf{0.74}    & 0.75    & 0.74    \\
& & Activation entropy        & 0.55 & 0.41 & 0.68 & 0.57    & 0.54    & 0.59    \\
& & Diff Activation map       & 0.53 & 0.50 & 0.56 & 0.54    & 0.52    & 0.54    \\
\hline\hline

\multirow{10}{*}{Vicuna-7B} & \multirow{4}{*}{Logit}      & Max token rank            & 0.55 & 0.34 & 0.74 & 0.66 & 0.67 & 0.65 \\
& & Min token prob            & 0.59 & 0.64 & 0.51 & 0.61 & 0.62 & 0.58 \\
& & Joint prob                & 0.63 & 0.72 & 0.45 & 0.67 & 0.57 & 0.72 \\
& & JSD                       & 0.53 & 0.53 & 0.52 & 0.53    & 0.53    & 0.55    \\
\cline{2-9}\cline{2-9}
& \multirow{3}{*}{Attention}  & Lookback ratio            & \textbf{0.82} & 0.83 & 0.84 & \textbf{0.87} & 0.81 & 0.88 \\
& & Attention sharpness       & 0.66 & 0.64 & 0.66 & 0.63 & 0.62 & 0.65 \\
& & Key token attention ratio & 0.83 & 0.81 & 0.83 & 0.86    & 0.84    & 0.85    \\
\cline{2-9}\cline{2-9}
& \multirow{3}{*}{Activation} & Hidden state              & \textbf{0.76} & 0.76 & 0.77 & \textbf{0.78}    & 0.79    & 0.80    \\
& & Activation entropy        & 0.56 & 0.43 & 0.63 & 0.57    & 0.61    & 0.56    \\
& & Diff Activation map       & 0.55 & 0.52 & 0.56 & 0.54    & 0.53    & 0.56    \\

\hline\hline
\end{tabular}
\caption{
We did an ablation study to show the ability of different features to classify hallucinations with facts.
}
\label{tab:ablation}
%\vspace{-15pt}
\end{table*}

\begin{table}[]
\centering
\begin{tabular}{c|c|c|c|c}
\hline\hline
Model & Dataset & CNNDM & HaluEval & NQ   \\
\hline\hline
\multirow{3}{*}{Llama-2-7B} & Fact. & 492   & 7632     & 1627 \\
& Hall. & 508   & 2368     & 1028 \\
& Sum.  & 1000  & 10000    & 2655 \\
\hline
\multirow{3}{*}{Vicuna-7B} & Fact. & 413   & 6171     & 1358 \\
& Hall. & 587   & 3829     & 1297 \\
& Sum.  & 1000  & 10000    & 2655 \\
\hline\hline
\end{tabular}
\caption{Target LLM's prediction towards 3 datasets.}
\label{tab:datasets}
\end{table}

To evaluate the performance of \sysname in detecting hallucinations, we utilized three datasets from distinct domains: CNN/Daily Mail (CNNDM), Natural Questions (NQ), and HaluEval. These datasets provide diverse scenarios, including summarization, question answering, and factuality evaluation, ensuring comprehensive testing of the proposed framework.

\sssec{CNN/Daily Mail (CNNDM)} \cite{hermann2015teaching}. This dataset is widely used for text summarization tasks. It consists of over 300,000 unique news articles from CNN and the Daily Mail, each paired with human-generated abstractive summaries. The dataset is divided into 286,817 training pairs, 13,368 validation pairs, and 11,487 test pairs. The average length of source documents in the training set is 766 words, while the summaries average 53 words. Hallucinations in this context are identified when generated summaries include content inconsistent with the original articles, emphasizing the model's ability to maintain fidelity in summarization tasks.

\sssec{Natural Questions (NQ)} \cite{kwiatkowski2019natural}. This question-answering dataset contains real user queries issued to the Google search engine, paired with corresponding Wikipedia pages. Each example includes a long answer (a paragraph) and, if applicable, a short answer (one or more entities) annotated by human annotators. The dataset comprises 307,373 training examples, 7,830 development examples, and 7,842 test examples. Hallucinations in NQ are defined as generated answers deviating from or contradicting the provided ground-truth answers, focusing on the factual correctness of responses.

\sssec{HaluEval} \cite{li2023halueval}. A specialized dataset designed to evaluate the factuality of language model outputs. It includes 5,000 general user queries with ChatGPT responses and 30,000 task-specific examples from three tasks: question answering, knowledge-grounded dialogue, and text summarization. For general user queries, the dataset adopts the 52K instruction tuning dataset from Alpaca. This dataset serves as a benchmark for testing both factual and hallucinated responses across multiple scenarios, offering a comprehensive evaluation of the model's detection capabilities.

%\subsection{Comparison with Previous Works}\label{sec:detect:sota}

%\subsection{Metrics}\label{sec:detect:metrics}

\begin{figure}
\centering
     \subfloat{\includegraphics[width=0.48\textwidth]{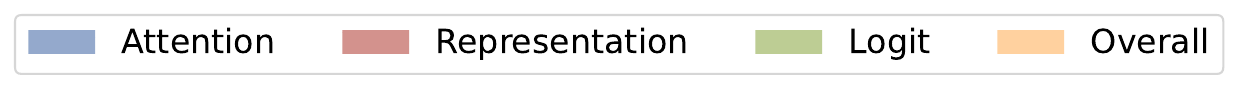}}
      \\
    \addtocounter{subfigure}{-1}
      \subfloat[Llama-2-7B]{\includegraphics[width=0.5\columnwidth]{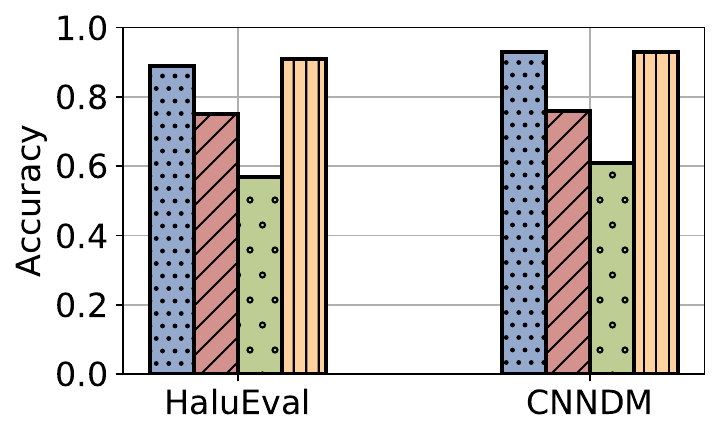}}
      \subfloat[Vicuna-7B]{\includegraphics[width=0.5\columnwidth]{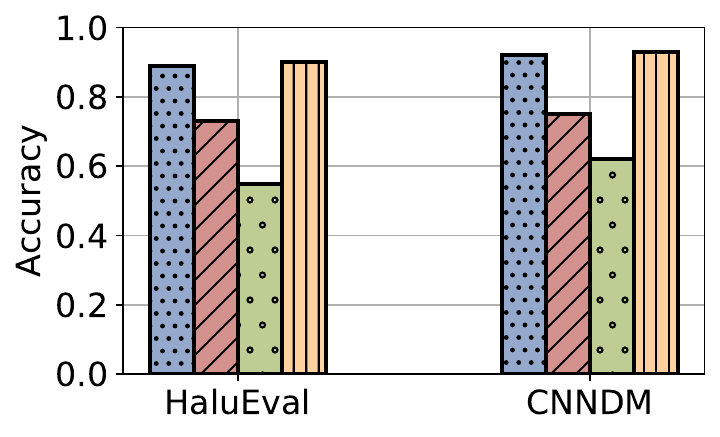}}
      \caption{
      We compare different groups of features.
      }
    \label{fig:detect_overall}
\end{figure}

\subsection{Ablation Study on Different Features}\label{sec:detect:ablation}

\sssec{Method}.
To analyze the effectiveness of various features in distinguishing hallucinated and factual outputs, we conducted an ablation study. In this experiment, we isolated each feature group (logit, attention, and activation) as the sole input to the detection model and evaluated their classification performance on the HaluEval and CNNDM datasets. Specifically, accuracy, recall for hallucinated outputs (\textit{Recall(Halu)}), and recall for factual outputs (\textit{Recall(Fact)}) were measured to assess the contribution of each feature. This approach enables a detailed comparison of feature importance and their respective strengths in different scenarios.

\sssec{Results}.
From Table~\ref{tab:ablation}, several observations can be made. Among the logit features, \textit{Joint Probability} achieves the highest \textit{Recall(Halu)} (0.72) on HaluEval, indicating its effectiveness in detecting hallucinations. However, its \textit{Recall(Fact)} is relatively low (0.37), making it less reliable for balanced detection. In contrast, \textit{Lookback Ratio}, an attention-based feature, performs consistently well across both datasets, achieving high accuracy (0.75 on HaluEval and 0.83 on CNNDM) and balanced recall values, suggesting its robustness in both hallucination and factual scenarios. Activation-based features, particularly the \textit{Hidden State}, show strong performance on HaluEval (accuracy of 0.76 and balanced recall values), but their efficacy diminishes on CNNDM, likely due to dataset-specific differences. Interestingly, some features like \textit{Attention Sharpness} and \textit{Activation Entropy}, while insightful during the understanding phase, demonstrate limited utility in actual classification, reflecting their inability to effectively capture discriminative information. These results suggest that while logit features excel in identifying hallucinations, attention-based features provide more consistent performance across datasets, making them suitable for general scenarios.

\subsection{Comparison between Different Token Selection Strategies}\label{sec:detect:token}

\begin{table}[]
\centering
\begin{tabular}{|M{1.5cm}|l|M{1.5cm}|M{1.5cm}|}
\hline\hline
Model & Strategy & CNNDM & HaluEval \\
\hline\hline
\multirow{7}{*}{Llama-2-7B} & All token           & 0.52     & 0.51     \\
& 1st token           & 0.51     & 0.53     \\
& Last token          & 0.53     & 0.51     \\
& Per token           & 0.68     & 0.74     \\
& Window(2, 1) & 0.78     & 0.81     \\
& Window(4, 2) & \textbf{0.85}     & \textbf{0.87}     \\
& Window(8, 4) & 0.83     & 0.82    \\
\hline
\multirow{7}{*}{Vicuna-7B} & All token           & 0.52     & 0.51     \\
& 1st token           & 0.50     & 0.52     \\
& Last token          & 0.52     & 0.52     \\
& Per token           & 0.71     & 0.75     \\
& Window(2, 1) & 0.80     & 0.82     \\
& Window(4, 2) & \textbf{0.87}     & \textbf{0.89}     \\
& Window(8, 4) & 0.85     & 0.83    \\
\hline\hline
\end{tabular}
\caption{Comparison between different token selection strategies.}
\label{tab:token}
\end{table}

% \usepackage{multirow}
\begin{table*}[]
\normalsize
\centering
\begin{tabular}{|M{1.6cm}|M{2cm}|M{3cm}|M{1cm}M{1cm}M{1cm}|M{1cm}M{1cm}M{1cm}|}
\hline\hline
\multirow{2}{*}{Model} &\multirow{2}{*}{Group} &
  \multirow{2}{*}{Feature} &
  \multicolumn{3}{c|}{CNNDM} &
  \multicolumn{3}{c|}{HaluEval} \\
  \cline{4-9}
 &&&
  \multicolumn{1}{c}{CNNDM} &
  \multicolumn{1}{c}{HaluEval} &
  \multicolumn{1}{c|}{NQ} &
  \multicolumn{1}{c}{CNNDM} &
  \multicolumn{1}{c}{HaluEval} &
  \multicolumn{1}{c|}{NQ} \\
  \hline\hline
\multirow{10}{*}{Llama-2-7B} & \multirow{4}{*}{Logit}      & Max token rank         & 0.6  & 0.52 & 0.58 & 0.51 & 0.53 & 0.52 \\
& & Min token prob         & 0.58 & 0.52 & 0.55 & 0.50 & 0.54 & 0.51 \\
& & Joint prob             & 0.63 & 0.56 & 0.61 & 0.54 & 0.54 & 0.52 \\
& & JSD                    & 0.51 & 0.51 & 0.50 & 0.50 & 0.51 & 0.51 \\
\cline{2-9}
& \multirow{3}{*}{Attention}  & Lookback ratio         & 0.83 & 0.54 & 0.66 & 0.57 & 0.75 & 0.50 \\
& & Attention sharpness    & 0.61 & 0.51 & 0.54 & 0.52 & 0.60 & 0.50 \\
& & Key token atten. ratio & 0.72 & 0.52 & 0.62 & 0.55 & 0.70 & 0.51 \\
\cline{2-9}
& \multirow{3}{*}{Activation} & Hidden state           & 0.74 & 0.53 & 0.55 & 0.52 & 0.76 & 0.50 \\
& & Activation entropy     & 0.57 & 0.52 & 0.51 & 0.51 & 0.55 & 0.52 \\
& & Activation map         & 0.54 & 0.52 & 0.50 & 0.52 & 0.53 & 0.50 \\
\hline \hline
\multirow{10}{*}{Vicuna-7B} & \multirow{4}{*}{Logit}      & Max token rank         & 0.6  & 0.52 & 0.58 & 0.51 & 0.53 & 0.52 \\
& & Min token prob         & 0.58 & 0.52 & 0.55 & 0.50 & 0.54 & 0.51 \\
& & Joint prob             & 0.63 & 0.56 & 0.61 & 0.54 & 0.54 & 0.52 \\
& & JSD                    & 0.51 & 0.51 & 0.50 & 0.50 & 0.51 & 0.51 \\
\cline{2-9}
& \multirow{3}{*}{Attention}  & Lookback ratio         & 0.83 & 0.54 & 0.66 & 0.57 & 0.75 & 0.50 \\
& & Attention sharpness    & 0.61 & 0.51 & 0.54 & 0.52 & 0.60 & 0.50 \\
& & Key token atten. ratio & 0.72 & 0.52 & 0.62 & 0.55 & 0.70 & 0.51 \\
\cline{2-9}
& \multirow{3}{*}{Activation} & Hidden state           & 0.74 & 0.53 & 0.55 & 0.52 & 0.76 & 0.50 \\
& & Activation entropy     & 0.57 & 0.52 & 0.51 & 0.51 & 0.55 & 0.52 \\
& & Activation map         & 0.54 & 0.52 & 0.50 & 0.52 & 0.53 & 0.50 \\
\hline \hline
\end{tabular}
%\vspace{-10pt}
\caption{We did experiment by using different datasets as train dataset and test dataset to see different internal state features' transferibility.}
\label{tab:transfer}
%\vspace{-10pt}
\end{table*}

\begin{table*}[ht]
\centering

\begin{tabular}{|l|l|l|l|c|}
\hline\hline
\textbf{Category} & \textbf{Feature Name}       & \textbf{Theo. Storage} & \textbf{Theo. Comp} & \textbf{Comp. Time Per Token (/s)} \\ \hline\hline
\multirow{2}{*}{Attention} 
                  & Attention Lookback Ratio    & $O(w \cdot H \cdot L)$               & $O(w \cdot H \cdot L)$                   & 3.27                                \\ \cline{2-5}
                  & Attention Allocation Sharpness & $O(w \cdot H \cdot L)$            & $O(w \cdot H \cdot L \cdot \log w)$      & 0.19                                \\ \hline\hline
\multirow{3}{*}{Activation}
                  & Last Layer Hidden State     & $O(w \cdot d)$                       & $O(w \cdot d)$                           & 0.09                                \\ \cline{2-5}
                  & Activation Map              & $O(w \cdot d \cdot m)$               & $O(w \cdot d \cdot m)$                   & 0.39                                \\ \cline{2-5}
                  & Activation Entropy          & $O(w \cdot m)$                       & $O(w \cdot m \cdot \log m)$              & 0.29                                \\ \hline\hline
\multirow{3}{*}{Logit}
                  & Min Token Probabilities     & $O(w \cdot L)$                       & $O(w \cdot L)$                           & 1.02                                \\ \cline{2-5}
                  & Max Token Ranks             & $O(w \cdot L)$                       & $O(w \cdot L \cdot \log w)$              & 1.15                               \\ \cline{2-5}
                  & Joint Token Probabilities   & $O(w \cdot L)$                       & $O(w \cdot L \cdot w)$                   & 0.73                                \\ \hline\hline
\end{tabular}
%\vspace{-10pt}
\caption{System Overhead for Different Features (Sliced Window, Window Size = 8)}
\label{tab:system_overhead}
%\vspace{-15pt}
\end{table*}

\sssec{Method}. To evaluate the impact of different token selection strategies on detection performance, we conducted experiments using various strategies, including \textit{All Token}, \textit{First Token}, \textit{Last Token}, \textit{Per Token}, and \textit{Sliced Window} with different window sizes. For each strategy, all extracted features were used as inputs to the detection model. The metrics reported in Table~\ref{tab:token} include accuracy scores on both the CNNDM and HaluEval datasets, allowing for a comprehensive comparison of the effectiveness of different token selection methods.

\sssec{Results}.
The results in Table~\ref{tab:token} reveal that the \textit{Sliced Window} strategy consistently outperforms others, achieving the highest accuracy (e.g., 0.87 on HaluEval with \textit{Window(4, 2)}), due to its ability to maintain uniform token input sizes across samples. This ensures that the detection model receives structured and balanced information. On the other hand, the \textit{Per Token} strategy, while intuitive, performs worse (e.g., 0.68 on CNNDM) likely due to noisy labeling issues rather than inherent flaws in the method itself. The \textit{First Token} and \textit{Last Token} strategies exhibit the lowest performance, as they fail to capture sufficient context for accurate classification. These findings suggest that while sliced window methods provide a structured and balanced approach suitable for diverse scenarios, the effectiveness of \textit{Per Token} could be improved with more refined labeling processes.

\subsection{Transferbility Study}\label{sec:detect:transfer}

\sssec{Methd}.
To evaluate the transferability of different internal state features, we conducted experiments using CNNDM and HaluEval as training datasets and tested their performance on CNNDM, HaluEval, and NQ, a benchmark dataset designed for question answering with a focus on fidelity-related hallucinations. For each experiment, we trained models using one dataset and evaluated the effectiveness of individual features on the remaining datasets to assess their cross-dataset generalizability.

\sssec{Results}.
The results in Table~\ref{tab:transfer} highlight several key findings. First, certain features trained on CNNDM, such as \textit{Lookback Ratio} and \textit{Joint Probability}, exhibit strong transferability to NQ, achieving relatively high accuracy scores (e.g., 0.66 and 0.61, respectively). This suggests that these features capture generalizable patterns that extend beyond the CNNDM dataset. However, features trained on HaluEval demonstrate poor transferability to CNNDM and NQ, with a significant drop in accuracy (e.g., \textit{Key Token Attention Ratio} and \textit{Hidden State} dropping to around 0.50). This discrepancy indicates that the features extracted from HaluEval are highly dataset-specific and may reflect idiosyncratic patterns rather than generalizable characteristics.

The lack of transferability between CNNDM and HaluEval is particularly notable. This suggests that the two datasets encode fundamentally different patterns of hallucination and factuality, potentially due to variations in text length, domain, or the nature of the hallucination tasks. These results highlight the importance of selecting datasets that align closely with the target domain when designing hallucination detection models and the challenges of building universally transferable feature-based systems.

\subsection{System Overhead}\label{sec:detect:cost}

\sssec{Method}.
To evaluate the computational efficiency and storage requirements of different features, we measured the theoretical storage overhead, theoretical computational complexity, and actual computation time under the sliced window setting with a window size of 8. The theoretical metrics are expressed in terms of the number of tokens in the window ($w$), the number of attention heads ($H$), the number of layers ($L$), the dimensionality of hidden states ($d$), and the intermediate dimensionality in feed-forward layers ($m$). These measurements allow us to quantify the resource demands for extracting each feature and to identify potential bottlenecks in real-world applications.

\sssec{Results}.
The theoretical results, as shown in Table~\ref{tab:system_overhead}, provide a detailed breakdown of storage and computation costs for features from attention, activation, and logit states. Actual computation time will be discussed once experimental data is finalized.

%\subsection{Different Target Model}\label{sec:detec:tar_models}
%\input{sections/6-attribution}
\section{Related Work}\label{sec:related}

In this section, we summarize current LLM hallucination detection (\S\ref{sec:related:detect}) and mitigation methods (\S\ref{sec:related:mitigate}), and categorize them based on internal ways and external ways.

\subsection{LLM Hallucinations Detection}\label{sec:related:detect}

Some previous works aim to detect hallucinated responses during LLM inference.
We divided them into 2 types: external and internal.

\sssec{External}.
Some works detect the hallucinations from the external response.
\cite{niu2024ragtruth, su2024mitigating, hu2024lrp4ragdetecting} utilizes RAG or KG to find the closest response from a ground-truth dataset, and compare it with the response from LLM to detect potential hallucinations.
Considering the high computing and storage cost, some lightweight external methods have also been proposed:
\cite{chen2024inside} propose new EigenScore metrics to evaluate responses' self-consistency.
\cite{zhang2023enhancing, sadat2023delucionqa} check the keywords within the response to check factuality.
\cite{kossen2024semantic} use semantics entropy from multiple queries to detect.

\sssec{Internal}.
Recently, researchers have tried to discover hallucination pattern within LLM inference process.
Most works directly use hidden states\cite{beigi2024internalinspector, chen2024inside, su2024unsupervised, duan2024llms, azaria2023internal} as the classification features, while activation state is also taken into consideration \cite{chen2024context, ji2024llm, he2024llm}.
Some works \cite{chuang2024lookback, yuksekgonul2023attention} also use attention as the classification feature since they have better interpretability.
Some researchers discover that the hallucinated tokens have lower token probability, and use it as a feature \cite{quevedo2024detecting, he2024llm}.

\subsection{LLM Hallucinations Mitigation}\label{sec:related:mitigate}

Previous works have proposed several methods to mitigate hallucinations in LLMs.
We divided previous mitigation methods into 3 types: external, model-based, and internal.

\sssec{External}.
Most works try to use external knowledge or inference paradigms to mitigate.
The most direct methods utilize outside knowledge to enhance LLMs' ability to generate factual responses, which can be categorized into knowledge graph-based \cite{sun2024thinkongraph, luo2024reasoning}, and RAG-based \cite{li2024chainofknowledge, tian2024finetuning}.
LLMs' ability to self-debug their response is also used to mitigate hallucinations \cite{chen2024teaching, gou2024critic}.
Some works also optimize the inference process of LLM, like dividing the task into sub-tasks \cite{pan2023fact}, building a chain for more robust reasoning \cite{luo2024reasoning, li2024chainofknowledge}.
Multi-agent paradigm is also considered in hallucination mitigation \cite{hong2024metagpt}, since the debate across multiple agents can enhance the response from a single agent.

\sssec{Model}.
Some works also aim at modifying models' weights for mitigation.
\cite{liu2024mitigating, li2024chainofknowledge, tian2024finetuning} addresses this issue by introducing a new dataset to fine-tune the model.
Some works aim to use full parameter fine-tuning \cite{tian2024finetuning}, while other works aim to use LoRA for more lightweight fine-tuning \cite{liu2024mitigating, li2024chainofknowledge}.
Besides fine-tuning, model editing is also a popular model-level hallucination mitigation method \cite{meng2023locating}, where developers or admins can only modify a few neural connections within the MLP layer to achieve accurate fact modification.

\sssec{Internal}.
Considering previous external methods and model methods' drawbacks on high cost and potential threat to catastrophic forgetting, some internal methods have been proposed.
Internal methods mainly focus on lightly modifying the decoding process of LLM without changing model weight.
\cite{azaria2023internal, chen2024context} discover potential mode within hidden state, and utilize it to mitigate hallucination.
\cite{chuang2024lookback} finds the significant difference between factual response and hallucinated response within attention scores, and intervenes attention score in the decoding process.
\cite{liu2024litcab} propose to use a new layer to intervene in logit output for better response.
\vspace{-10pt}
\section{Limitation \& Discussion}\label{sec:discussion}

%\sssec{High system cost}.
%The high computational and storage cost associated with many features significantly limits their practicality. Features such as \textit{Activation Map} and \textit{Joint Token Probabilities}, while effective in certain scenarios, require substantial resources due to their high theoretical complexity. This can lead to increased inference latency, making them unsuitable for real-time or resource-constrained applications.

\sssec{Limited transferability}.
Based on the transferability experiments and insights from the RAG analysis, it is evident that the internal features of LLMs differ greatly across different scenarios. This lack of consistency results in poor cross-dataset generalizability. For example, features trained on CNNDM fail to perform well on HaluEval and vice versa, highlighting the dataset-specific nature of these features. This presents a significant challenge in developing universally applicable detection systems.

\sssec{Potential explainability}.
Despite the limitations, the proposed approach offers a promising avenue for explainability. Attention-based features, such as \textit{Lookback Ratio}, and logit-based features, such as \textit{Joint Token Probabilities}, provide interpretable insights into the model's reasoning process. These features allow researchers to better understand why certain outputs are classified as hallucinated, thereby enhancing trust in the system and opening up new possibilities for debugging and model refinement.
\vspace{-10pt}
\section{Conclusion}\label{sec:conclusion}

In this work, we proposed and evaluated a comprehensive framework for understanding and detecting hallucinations in large language models through internal state analysis. Our experiments demonstrated the potential of using attention, activation, and logit-based features to distinguish between hallucinated and factual outputs. However, challenges remain, including high computational costs, limited transferability of features across datasets, and the need for more efficient and generalizable detection strategies. Despite these limitations, the approach shows promise for improving the explainability of hallucination detection, especially through interpretable features like attention and logits. Future work will focus on optimizing feature extraction for real-time applications and exploring methods to enhance feature transferability across diverse scenarios.

\newpage\clearpage

\section*{Ethics Consideration}

This study does not involve significant ethical concerns. The research is conducted using publicly available datasets (e.g., CNNDM, HaluEval), ensuring compliance with their usage policies and avoiding using private or sensitive data. No live systems were tested, and the methodology avoids actions that could harm users or disrupt services. Furthermore, the study emphasizes responsible reporting and discusses safeguards to prevent potential misuse of the findings. Ethical principles, including beneficence, justice, and respect for privacy, are inherently upheld in this research.

\section*{Open Science}

To promote transparency and reproducibility in research, we commit to sharing the source code, datasets, and implementation details of our proposed framework, \sysname. Upon acceptance, all relevant artifacts will be made publicly available on a trusted repository (e.g., GitHub or Zenodo) under an open-source license. This includes:
\begin{enumerate}
    \item The complete codebase for \sysname, covering inner state extraction, feature computation, and hallucination detection.
    \item Instructions for reproducing the experiments, including the preprocessing of datasets and evaluation metrics.
    \item A detailed README file to guide users in replicating the results.
\end{enumerate}

We believe that open science fosters collaboration and innovation, and we aim to contribute to the broader research community by ensuring the accessibility and usability of our work.

\clearpage
\newpage
{\footnotesize \bibliographystyle{acm}
\bibliography{sample}}

\begin{thebibliography}{10}

\bibitem{azaria2023internal}
{\sc Azaria, A., and Mitchell, T.}
\newblock The internal state of an llm knows when it's lying.
\newblock {\em arXiv preprint arXiv:2304.13734\/} (2023).

\bibitem{beigi2024internalinspector}
{\sc Beigi, M., Shen, Y., Yang, R., Lin, Z., Wang, Q., Mohan, A., He, J., Jin, M., Lu, C.-T., and Huang, L.}
\newblock Internalinspector: Robust confidence estimation in llms through internal states.
\newblock {\em arXiv preprint arXiv:2406.12053\/} (2024).

\bibitem{belyi2024luna}
{\sc Belyi, M., Friel, R., Shao, S., and Sanyal, A.}
\newblock Luna: An evaluation foundation model to catch language model hallucinations with high accuracy and low cost.
\newblock {\em arXiv preprint arXiv:2406.00975\/} (2024).

\bibitem{chefer2021transformer}
{\sc Chefer, H., Gur, S., and Wolf, L.}
\newblock Transformer interpretability beyond attention visualization.
\newblock In {\em Proceedings of the IEEE/CVF conference on computer vision and pattern recognition\/} (2021), pp.~782--791.

\bibitem{chen2024inside}
{\sc Chen, C., Liu, K., Chen, Z., Gu, Y., Wu, Y., Tao, M., Fu, Z., and Ye, J.}
\newblock Inside: Llms' internal states retain the power of hallucination detection.
\newblock {\em arXiv preprint arXiv:2402.03744\/} (2024).

\bibitem{chen2024context}
{\sc Chen, S., Xiong, M., Liu, J., Wu, Z., Xiao, T., Gao, S., and He, J.}
\newblock In-context sharpness as alerts: An inner representation perspective for hallucination mitigation.
\newblock {\em arXiv preprint arXiv:2403.01548\/} (2024).

\bibitem{chen2024teaching}
{\sc Chen, X., Lin, M., Sch{\"a}rli, N., and Zhou, D.}
\newblock Teaching large language models to self-debug.
\newblock In {\em The Twelfth International Conference on Learning Representations\/} (2024).

\bibitem{chuang2024lookback}
{\sc Chuang, Y.-S., Qiu, L., Hsieh, C.-Y., Krishna, R., Kim, Y., and Glass, J.}
\newblock Lookback lens: Detecting and mitigating contextual hallucinations in large language models using only attention maps.
\newblock {\em arXiv preprint arXiv:2407.07071\/} (2024).

\bibitem{cui2023chatlaw}
{\sc Cui, J., Li, Z., Yan, Y., Chen, B., and Yuan, L.}
\newblock Chatlaw: Open-source legal large language model with integrated external knowledge bases.
\newblock {\em CoRR\/} (2023).

\bibitem{dar2022analyzing}
{\sc Dar, G., Geva, M., Gupta, A., and Berant, J.}
\newblock Analyzing transformers in embedding space.
\newblock {\em arXiv preprint arXiv:2209.02535\/} (2022).

\bibitem{duan2024llms}
{\sc Duan, H., Yang, Y., and Tam, K.~Y.}
\newblock Do llms know about hallucination? an empirical investigation of llm's hidden states.
\newblock {\em arXiv preprint arXiv:2402.09733\/} (2024).

\bibitem{gao2023retrieval}
{\sc Gao, Y., Xiong, Y., Gao, X., Jia, K., Pan, J., Bi, Y., Dai, Y., Sun, J., Wang, M., and Wang, H.}
\newblock Retrieval-augmented generation for large language models: A survey.
\newblock {\em arXiv preprint arXiv:2312.10997\/} (2023).

\bibitem{gou2024critic}
{\sc Gou, Z., Shao, Z., Gong, Y., yelong shen, Yang, Y., Duan, N., and Chen, W.}
\newblock {CRITIC}: Large language models can self-correct with tool-interactive critiquing.
\newblock In {\em The Twelfth International Conference on Learning Representations\/} (2024).

\bibitem{guha2024legalbench}
{\sc Guha, N., Nyarko, J., Ho, D., R{\'e}, C., Chilton, A., Chohlas-Wood, A., Peters, A., Waldon, B., Rockmore, D., Zambrano, D., et~al.}
\newblock Legalbench: A collaboratively built benchmark for measuring legal reasoning in large language models.
\newblock {\em Advances in Neural Information Processing Systems 36\/} (2024).

\bibitem{hanna2023comparative}
{\sc Hanna, E., and Levic, A.}
\newblock Comparative analysis of language models: hallucinations in chatgpt: Prompt study, 2023.

\bibitem{harrington2024mitigating}
{\sc Harrington, F., Rosenthal, E., and Swinburne, M.}
\newblock Mitigating hallucinations in large language models with sliding generation and self-checks.
\newblock {\em Authorea Preprints\/} (2024).

\bibitem{he2024llm}
{\sc He, J., Gong, Y., Lin, Z., Zhao, Y., Chen, K., et~al.}
\newblock Llm factoscope: Uncovering llms’ factual discernment through measuring inner states.
\newblock In {\em Findings of the Association for Computational Linguistics ACL 2024\/} (2024), pp.~10218--10230.

\bibitem{hermann2015teaching}
{\sc Hermann, K.~M., Kocisky, T., Grefenstette, E., Espeholt, L., Kay, W., Suleyman, M., and Blunsom, P.}
\newblock Teaching machines to read and comprehend.
\newblock {\em Advances in neural information processing systems 28\/} (2015).

\bibitem{hong2024metagpt}
{\sc Hong, S., Zhuge, M., Chen, J., Zheng, X., Cheng, Y., Wang, J., Zhang, C., Wang, Z., Yau, S. K.~S., Lin, Z., Zhou, L., Ran, C., Xiao, L., Wu, C., and Schmidhuber, J.}
\newblock Meta{GPT}: Meta programming for a multi-agent collaborative framework.
\newblock In {\em The Twelfth International Conference on Learning Representations\/} (2024).

\bibitem{hu2024lrp4ragdetecting}
{\sc Hu, H., Sun, Y., and Zhang, Q.}
\newblock Lrp4rag: Detecting hallucinations in retrieval-augmented generation via layer-wise relevance propagation, 2024.

\bibitem{huang2023survey}
{\sc Huang, L., Yu, W., Ma, W., Zhong, W., Feng, Z., Wang, H., Chen, Q., Peng, W., Feng, X., Qin, B., et~al.}
\newblock A survey on hallucination in large language models: Principles, taxonomy, challenges, and open questions.
\newblock {\em arXiv preprint arXiv:2311.05232\/} (2023).

\bibitem{ji2024llm}
{\sc Ji, Z., Chen, D., Ishii, E., Cahyawijaya, S., Bang, Y., Wilie, B., and Fung, P.}
\newblock Llm internal states reveal hallucination risk faced with a query.
\newblock {\em arXiv preprint arXiv:2407.03282\/} (2024).

\bibitem{ji2023survey}
{\sc Ji, Z., Lee, N., Frieske, R., Yu, T., Su, D., Xu, Y., Ishii, E., Bang, Y.~J., Madotto, A., and Fung, P.}
\newblock Survey of hallucination in natural language generation.
\newblock {\em ACM Computing Surveys 55}, 12 (2023), 1--38.

\bibitem{kossen2024semantic}
{\sc Kossen, J., Han, J., Razzak, M., Schut, L., Malik, S., and Gal, Y.}
\newblock Semantic entropy probes: Robust and cheap hallucination detection in llms, 2024.

\bibitem{kwiatkowski2019natural}
{\sc Kwiatkowski, T., Palomaki, J., Redfield, O., Collins, M., Parikh, A., Alberti, C., Epstein, D., Polosukhin, I., Devlin, J., Lee, K., et~al.}
\newblock Natural questions: a benchmark for question answering research.
\newblock {\em Transactions of the Association for Computational Linguistics 7\/} (2019), 453--466.

\bibitem{li2024dawn}
{\sc Li, J., Chen, J., Ren, R., Cheng, X., Zhao, W.~X., Nie, J.-Y., and Wen, J.-R.}
\newblock The dawn after the dark: An empirical study on factuality hallucination in large language models.
\newblock {\em arXiv preprint arXiv:2401.03205\/} (2024).

\bibitem{li2023halueval}
{\sc Li, J., Cheng, X., Zhao, W.~X., Nie, J.-Y., and Wen, J.-R.}
\newblock Halueval: A large-scale hallucination evaluation benchmark for large language models.
\newblock {\em arXiv preprint arXiv:2305.11747\/} (2023).

\bibitem{li2024enhancing}
{\sc Li, J., Yuan, Y., and Zhang, Z.}
\newblock Enhancing llm factual accuracy with rag to counter hallucinations: A case study on domain-specific queries in private knowledge-bases.
\newblock {\em arXiv preprint arXiv:2403.10446\/} (2024).

\bibitem{li2024chainofknowledge}
{\sc Li, X., Zhao, R., Chia, Y.~K., Ding, B., Joty, S., Poria, S., and Bing, L.}
\newblock Chain-of-knowledge: Grounding large language models via dynamic knowledge adapting over heterogeneous sources.
\newblock In {\em The Twelfth International Conference on Learning Representations\/} (2024).

\bibitem{liu2024mitigating}
{\sc Liu, F., Lin, K., Li, L., Wang, J., Yacoob, Y., and Wang, L.}
\newblock Mitigating hallucination in large multi-modal models via robust instruction tuning.
\newblock In {\em The Twelfth International Conference on Learning Representations\/} (2024).

\bibitem{liu2024litcab}
{\sc Liu, X., Khalifa, M., and Wang, L.}
\newblock Litcab: Lightweight language model calibration over short- and long-form responses.
\newblock In {\em The Twelfth International Conference on Learning Representations\/} (2024).

\bibitem{luo2024reasoning}
{\sc LUO, L., Li, Y.-F., Haf, R., and Pan, S.}
\newblock Reasoning on graphs: Faithful and interpretable large language model reasoning.
\newblock In {\em The Twelfth International Conference on Learning Representations\/} (2024).

\bibitem{manakul2023selfcheckgpt}
{\sc Manakul, P., Liusie, A., and Gales, M.~J.}
\newblock Selfcheckgpt: Zero-resource black-box hallucination detection for generative large language models.
\newblock {\em arXiv preprint arXiv:2303.08896\/} (2023).

\bibitem{mckenna2023sources}
{\sc McKenna, N., Li, T., Cheng, L., Hosseini, M.~J., Johnson, M., and Steedman, M.}
\newblock Sources of hallucination by large language models on inference tasks.
\newblock {\em arXiv preprint arXiv:2305.14552\/} (2023).

\bibitem{meng2023locating}
{\sc Meng, K., Bau, D., Andonian, A., and Belinkov, Y.}
\newblock Locating and editing factual associations in gpt, 2023.

\bibitem{mickus2022dissect}
{\sc Mickus, T., Paperno, D., and Constant, M.}
\newblock How to dissect a muppet: The structure of transformer embedding spaces.
\newblock {\em Transactions of the Association for Computational Linguistics 10\/} (2022), 981--996.

\bibitem{mundler2023self}
{\sc M{\"u}ndler, N., He, J., Jenko, S., and Vechev, M.}
\newblock Self-contradictory hallucinations of large language models: Evaluation, detection and mitigation.
\newblock {\em arXiv preprint arXiv:2305.15852\/} (2023).

\bibitem{niu2023ragtruth}
{\sc Niu, C., Wu, Y., Zhu, J., Xu, S., Shum, K., Zhong, R., Song, J., and Zhang, T.}
\newblock Ragtruth: A hallucination corpus for developing trustworthy retrieval-augmented language models.
\newblock {\em arXiv preprint arXiv:2401.00396\/} (2023).

\bibitem{niu2024ragtruth}
{\sc Niu, C., Wu, Y., Zhu, J., Xu, S., Shum, K., Zhong, R., Song, J., and Zhang, T.}
\newblock Ragtruth: A hallucination corpus for developing trustworthy retrieval-augmented language models, 2024.

\bibitem{pan2023fact}
{\sc Pan, L., Wu, X., Lu, X., Luu, A.~T., Wang, W.~Y., Kan, M.-Y., and Nakov, P.}
\newblock Fact-checking complex claims with program-guided reasoning, 2023.

\bibitem{peng2023study}
{\sc Peng, C., Yang, X., Chen, A., Smith, K.~E., PourNejatian, N., Costa, A.~B., Martin, C., Flores, M.~G., Zhang, Y., Magoc, T., et~al.}
\newblock A study of generative large language model for medical research and healthcare.
\newblock {\em NPJ digital medicine 6}, 1 (2023), 210.

\bibitem{quevedo2024detecting}
{\sc Quevedo, E., Yero, J., Koerner, R., Rivas, P., and Cerny, T.}
\newblock Detecting hallucinations in large language model generation: A token probability approach.
\newblock {\em arXiv preprint arXiv:2405.19648\/} (2024).

\bibitem{rawte2023survey}
{\sc Rawte, V., Sheth, A., and Das, A.}
\newblock A survey of hallucination in large foundation models.
\newblock {\em arXiv preprint arXiv:2309.05922\/} (2023).

\bibitem{rigotti2021attention}
{\sc Rigotti, M., Miksovic, C., Giurgiu, I., Gschwind, T., and Scotton, P.}
\newblock Attention-based interpretability with concept transformers.
\newblock In {\em International conference on learning representations\/} (2021).

\bibitem{sadat2023delucionqa}
{\sc Sadat, M., Zhou, Z., Lange, L., Araki, J., Gundroo, A., Wang, B., Menon, R.~R., Parvez, M.~R., and Feng, Z.}
\newblock Delucionqa: Detecting hallucinations in domain-specific question answering, 2023.

\bibitem{su2024mitigating}
{\sc Su, W., Tang, Y., Ai, Q., Wang, C., Wu, Z., and Liu, Y.}
\newblock Mitigating entity-level hallucination in large language models, 2024.

\bibitem{su2024unsupervised}
{\sc Su, W., Wang, C., Ai, Q., Hu, Y., Wu, Z., Zhou, Y., and Liu, Y.}
\newblock Unsupervised real-time hallucination detection based on the internal states of large language models.
\newblock {\em arXiv preprint arXiv:2403.06448\/} (2024).

\bibitem{sun2024thinkongraph}
{\sc Sun, J., Xu, C., Tang, L., Wang, S., Lin, C., Gong, Y., Ni, L., Shum, H.-Y., and Guo, J.}
\newblock Think-on-graph: Deep and responsible reasoning of large language model on knowledge graph.
\newblock In {\em The Twelfth International Conference on Learning Representations\/} (2024).

\bibitem{thirunavukarasu2023large}
{\sc Thirunavukarasu, A.~J., Ting, D. S.~J., Elangovan, K., Gutierrez, L., Tan, T.~F., and Ting, D. S.~W.}
\newblock Large language models in medicine.
\newblock {\em Nature medicine 29}, 8 (2023), 1930--1940.

\bibitem{tian2024finetuning}
{\sc Tian, K., Mitchell, E., Yao, H., Manning, C.~D., and Finn, C.}
\newblock Fine-tuning language models for factuality.
\newblock In {\em The Twelfth International Conference on Learning Representations\/} (2024).

\bibitem{tonmoy2024comprehensive}
{\sc Tonmoy, S., Zaman, S., Jain, V., Rani, A., Rawte, V., Chadha, A., and Das, A.}
\newblock A comprehensive survey of hallucination mitigation techniques in large language models.
\newblock {\em arXiv preprint arXiv:2401.01313\/} (2024).

\bibitem{vaswani2017attention}
{\sc Vaswani, A.}
\newblock Attention is all you need.
\newblock {\em Advances in Neural Information Processing Systems\/} (2017).

\bibitem{vig2019analyzing}
{\sc Vig, J., and Belinkov, Y.}
\newblock Analyzing the structure of attention in a transformer language model.
\newblock {\em arXiv preprint arXiv:1906.04284\/} (2019).

\bibitem{wu2023bloomberggpt}
{\sc Wu, S., Irsoy, O., Lu, S., Dabravolski, V., Dredze, M., Gehrmann, S., Kambadur, P., Rosenberg, D., and Mann, G.}
\newblock Bloomberggpt: A large language model for finance.
\newblock {\em arXiv preprint arXiv:2303.17564\/} (2023).

\bibitem{yang2022large}
{\sc Yang, X., Chen, A., PourNejatian, N., Shin, H.~C., Smith, K.~E., Parisien, C., Compas, C., Martin, C., Costa, A.~B., Flores, M.~G., et~al.}
\newblock A large language model for electronic health records.
\newblock {\em NPJ digital medicine 5}, 1 (2022), 194.

\bibitem{yeh2023attentionviz}
{\sc Yeh, C., Chen, Y., Wu, A., Chen, C., Vi{\'e}gas, F., and Wattenberg, M.}
\newblock Attentionviz: A global view of transformer attention.
\newblock {\em IEEE Transactions on Visualization and Computer Graphics\/} (2023).

\bibitem{yuksekgonul2023attention}
{\sc Yuksekgonul, M., Chandrasekaran, V., Jones, E., Gunasekar, S., Naik, R., Palangi, H., Kamar, E., and Nushi, B.}
\newblock Attention satisfies: A constraint-satisfaction lens on factual errors of language models.
\newblock {\em arXiv preprint arXiv:2309.15098\/} (2023).

\bibitem{zhang2023enhancing}
{\sc Zhang, T., Qiu, L., Guo, Q., Deng, C., Zhang, Y., Zhang, Z., Zhou, C., Wang, X., and Fu, L.}
\newblock Enhancing uncertainty-based hallucination detection with stronger focus, 2023.

\bibitem{zhang2023siren}
{\sc Zhang, Y., Li, Y., Cui, L., Cai, D., Liu, L., Fu, T., Huang, X., Zhao, E., Zhang, Y., Chen, Y., et~al.}
\newblock Siren's song in the ai ocean: a survey on hallucination in large language models.
\newblock {\em arXiv preprint arXiv:2309.01219\/} (2023).

\bibitem{zhao2024explainability}
{\sc Zhao, H., Chen, H., Yang, F., Liu, N., Deng, H., Cai, H., Wang, S., Yin, D., and Du, M.}
\newblock Explainability for large language models: A survey.
\newblock {\em ACM Transactions on Intelligent Systems and Technology 15}, 2 (2024), 1--38.

\end{thebibliography}

\clearpage
\newpage
\appendix
\onecolumn

\section{Prompt templates}

\subsection{Hallucination judgement prompt}

\begin{tcolorbox}[title=Prompt Template of Labeling Process for Faithful Hallucination Benchmark, colback=gray!20, colframe=gray!75, rounded corners, sharp corners=northeast, sharp corners=southwest]
\small

You will be provided with a document and a proposed summary. Your task is to determine if the proposed summary can be directly inferred from the document. If the summary contains any information not found in the document, it is considered false. Even if the summary is different from a ground truth summary, it might still be true, as long as it doesn't contain false information. \\

For each proposed summary, explain why it is true or false based on the information from the document. Focus only on the original document's content, disregarding any external context. \\

After your explanation, give your final conclusion as **Conclusion: True** if the proposed summary is completely accurate based on the document, or **Conclusion: False** if it contains any incorrect or unsupported information. If your conclusion is 'False', identify the exact phrases or name entities from the summary that is incorrect by stating **Problematic Spans: [the inaccurate text spans from the summary, in Python list of strings format]**. \\

\# Document \#: \textcolor{orange}{\$document} \\

\# Ground Truth Summary \#: \textcolor{blue}{\$gt\_response} \\

\# Proposed Summary \#: \textcolor{red}{\$response} \\

Write your explanation first, and then give your final conclusion as **Conclusion: True** if the proposed summary is completely accurate based on the document, or **Conclusion: False** if it contains any incorrect or unsupported information. Add **Problematic Spans: [the exact inaccurate text spans from the summary, in a list of strings]** if your conclusion is 'False'.

\end{tcolorbox}

\textcolor{orange}{\$document} is the original document text for the faithful hallucinated question. \textcolor{blue}{\$gt\_response} is the ground-truth response for the question. \textcolor{red}{\$response} is target LLM's response towards the question.

\begin{tcolorbox}[title=Prompt Template of Labeling Process for Factual Hallucination Benchmark, colback=gray!20, colframe=gray!75, rounded corners, sharp corners=northeast, sharp corners=southwest]
\small

You will be provided with a question (along with the groundtruth answer) and a proposed answer. Your task is to determine if the proposed answer align with the groundtruth answer. If the answer is not aligned with the groundtruth answer, it is considered false. Even if the answer is not totally the same with ground truth answer, it might still be true (For example, it is just a difference in expression or an alias for some nouns.). \\

For each proposed answer, explain why it is true or false based on the given question and the groundtruth answer. \\

After your explanation, give your final conclusion as **Conclusion: True** if the proposed answer is completely accurate, or **Conclusion: False** if it contains any incorrect or unsupported information. If your conclusion is 'False', identify the exact phrases or name entities from the answer that is incorrect by stating **Problematic Spans: [the inaccurate text spans from the answer, in Python list of strings format]**. It should only include the exact wrong phrases or name entities but not the full sentence. Avoid to include the whole sentence.\\

\# Question \#: \textcolor{orange}{\$question} \\

\# Ground Truth Summary \#: \textcolor{blue}{\$gt\_response} \\

\# Proposed Summary \#: \textcolor{red}{\$response} \\

Write your explanation first, and then give your final conclusion as **Conclusion: True** if the proposed answer is completely accurate, or **Conclusion: False** if it contains any incorrect or unsupported information. Add **Problematic Spans: [the exact inaccurate text spans from the answer, in a list of strings]** if your conclusion is 'False'.

\end{tcolorbox}

\textcolor{orange}{\$question} is the original question text for the factual hallucinated question. \textcolor{blue}{\$gt\_response} is the ground-truth response for the question. \textcolor{red}{\$response} is target LLM's response towards the question.

\end{document}